\pdfoutput=1
\documentclass[10pt, logo, twocolumn, copyright, nonumbering]{nvidiatechreport}

\usepackage{hyperref}
\usepackage{url}
\usepackage{colortbl}
\usepackage{nicefrac}       %
\usepackage{microtype}      %
\usepackage[dvipsnames]{xcolor}         %
\usepackage{multirow}
\usepackage{multicol}
\usepackage{graphicx}
\usepackage{xspace}
\usepackage{amsmath}
\usepackage{adjustbox}
\usepackage{tcolorbox}
\usepackage{amssymb}
\usepackage{enumitem}
\usepackage{wrapfig}
\usepackage{xcolor}
\usepackage{subcaption}    %
\usepackage{float}
\usepackage{stfloats}
\usepackage{amsmath}
\usepackage{listings}
\usepackage{mdframed}
\usepackage{tcolorbox}
\usepackage{arydshln}
\usepackage{booktabs} 
\usepackage{color,soul}
\usepackage{makecell}
\usepackage[numbers]{natbib}

\tcbuselibrary{listings,breakable}

\definecolor{pastelblue}{RGB}{173,216,230}
\definecolor{pastelyellow}{RGB}{255,253,208}
\definecolor{pastelpink}{RGB}{255,209,220}
\definecolor{pastelgreen}{RGB}{176,226,172}
\definecolor{pastellavender}{RGB}{230,230,250}

\definecolor{NvidiaGreen}{RGB}{118, 185, 0}
\sethlcolor{red!15}

\title{\centering{AIMO-2 Winning Solution: Building State-of-the-Art Mathematical Reasoning Models with OpenMathReasoning dataset}}

\author{
\centering{
Ivan Moshkov, Darragh Hanley, 
Ivan Sorokin,  Shubham Toshniwal,
Christof Henkel, \hspace{1.8in}
Benedikt Schifferer,
Wei Du,
Igor Gitman
}
}

\begin{abstract}
\textbf{Abstract:}

This paper presents our winning submission to the AI Mathematical Olympiad - Progress Prize 2 (AIMO-2) competition. 
Our recipe for building state-of-the-art mathematical reasoning models relies on three key pillars. 
First, we create a large-scale dataset comprising 540K unique high-quality math problems, including olympiad-level problems, and their 3.2M long-reasoning solutions. Second, we develop a novel method to integrate code execution with long reasoning models through iterative training, generation, and quality filtering, resulting in 1.7M high-quality Tool-Integrated Reasoning solutions. 
Third, we create a pipeline to train models to select the most promising solution from many candidates. We show that such generative solution selection (GenSelect) can significantly improve upon majority voting baseline. Combining these ideas, we train a series of models that achieve state-of-the-art results on mathematical reasoning benchmarks. To facilitate further research, we release our code, models, and the complete \texttt{OpenMathReasoning} dataset under a commercially permissive license.
\end{abstract}

\begin{document}

\maketitle

\section{Introduction}

\begin{figure*}[!b]
    \centering
    \includegraphics[width=\linewidth]{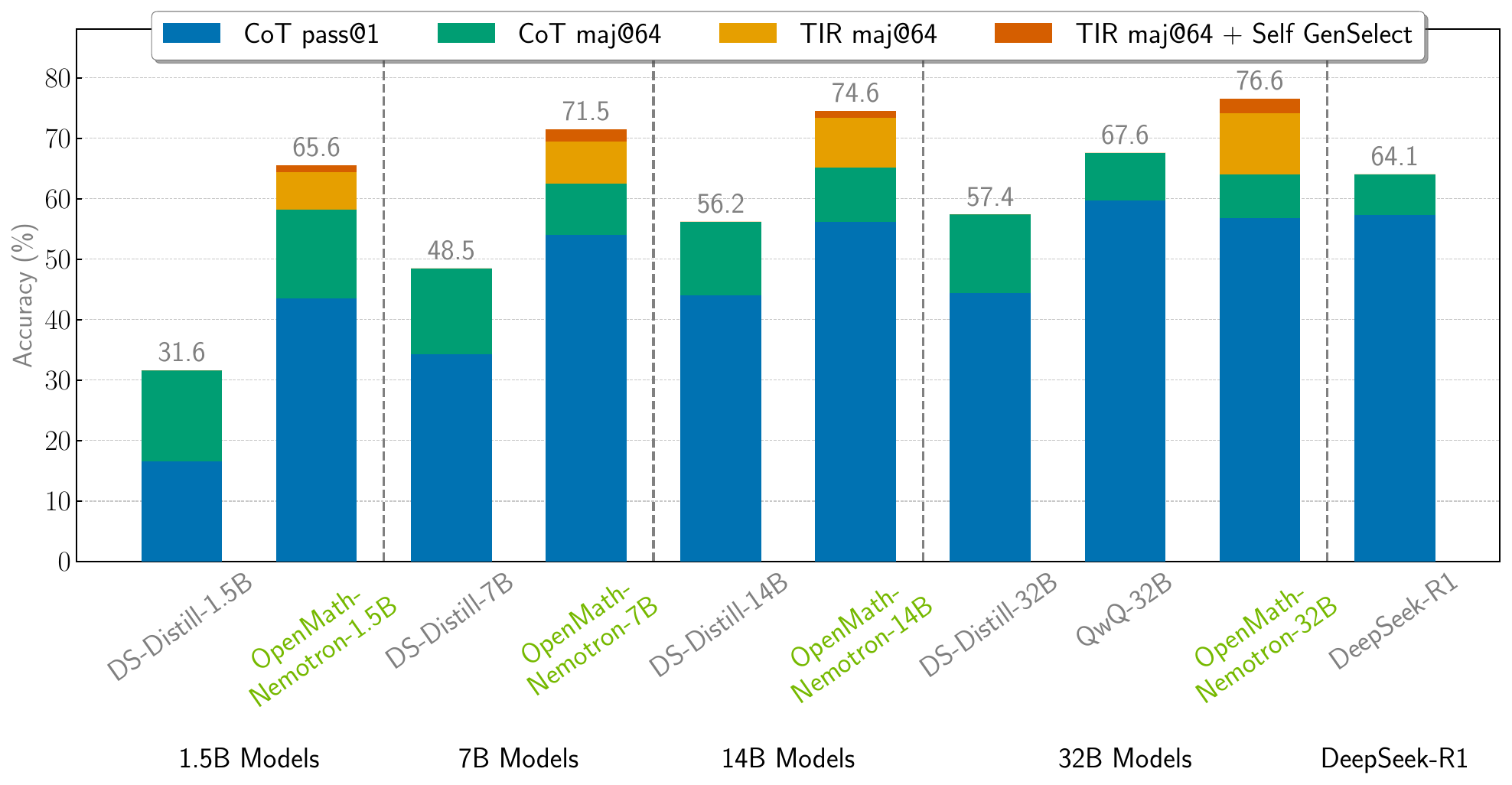}
    \captionsetup{justification=centering}
    \caption{Accuracy on math problems from AIME and HMMT competitions.}
    \label{fig:yourlabel}
\end{figure*}

Recent advances in large language models (LLMs) have significantly improved their ability to solve complex reasoning tasks, including olympiad-level mathematics. A key idea behind this progress has been to allow models to spend more tokens thinking about the solution before producing the final answer. Initially, models were trained to produce a series of intermediate solution steps (chain‑of‑thought (CoT)~\cite{wei2022chain}). More recently, \textit{long reasoning} models~\citep{jaech2024openai, guo2025deepseek} have learned to reflect on their work, exploring and refining multiple strategies within a single generation. This has led to further improvements across mathematics, coding, and scientific domains. To keep pace with this rapid development, the community has introduced increasingly challenging benchmarks and competitions that help to evaluate the progress.

The \href{https://www.kaggle.com/competitions/ai-mathematical-olympiad-progress-prize-2}{AI Mathematical Olympiad - Progress Prize 2 (AIMO-2)} is an initiative designed to assess advancements in this domain by challenging participants to create models capable of solving 50 difficult, national-level mathematical problems within strict computational limits. These problems were never published online, ensuring a more rigorous evaluation compared to traditional benchmarks. This report details our first-place submission to the competition, which correctly solved 34 out of 50 problems on the private test set. To develop the winning recipe, we focused on addressing several limitations of the publicly available reasoning models that we describe below.

\textbf{Large-scale long-reasoning dataset (\S\ref{sec:data}).} To improve existing models we started by collecting an extensive set of mathematical problems from the internet. We developed an LLM-based problem extraction and refinement pipeline to construct a dataset of 540K unique problems. Using this dataset, we then generated 3.2M long-reasoning CoT solutions by prompting \texttt{DeepSeek-R1}~\cite{guo2025deepseek} and \texttt{QwQ-32B}~\cite{qwq32b}. Training \texttt{Qwen2.5-Base} models~\cite{qwen2025qwen25technicalreport} on this large-scale distillation data, we are able to surpass the accuracy of all other open-weight models of comparable size, except for \texttt{QwQ-32B}, which is slightly better than our 32B model. 

\textbf{Tool-Integrated Reasoning (\S\ref{sec:tir}).} To improve the results further we developed a method for integrating code execution into long-reasoning generations. Our initial attempts to elicit Tool-Integrated Reasoning (TIR) from \texttt{DeepSeek-R1} and \texttt{QwQ-32B} through simple prompting proved unsuccessful. We hypothesize that these models struggle to deviate from their standard solution format due to extensive training on reasoning tasks and limited exposure to instruction-following. To overcome this challenge, we built a pipeline that starts with a small-scale reasoning finetuning of an \textit{instruction-following} model~\cite{ye2025limo}. By prompting this model to generate long-reasoning TIR solutions followed by aggressive quality filtering, we established an initial dataset suitable for training. Through multiple iterations of training, generation, and filtering, we constructed a 1.7M TIR solution set that was crucial for improving the accuracy of our final models. To make TIR more efficient, we also developed a method to accurately control the number of code executions the model is allowed to make for each generation.

\textbf{Generative Solution Selection (\S\ref{sec:genrm}).} A common approach to maximize model accuracy is to generate multiple candidate solutions and select the most promising one. While majority voting~\cite{DBLP:conf/iclr/0002WSLCNCZ23} serves as a strong baseline, its performance falls significantly short of the theoretical maximum performance of pass@k.   
To address this limitation, we developed a pipeline for training models to identify the most promising solution when presented with multiple candidates. We generated 566K selection examples to train our models. Although this approach showed considerable promise, we ultimately were unable to integrate it into our AIMO-2 Kaggle submission due to the competition's strict time constraints.

Combining these three innovations, we developed a series of state-of-the-art open-weight math reasoning models with 1.5B, 7B, 14B, and 32B parameters. Each model supports CoT, TIR and GenSelect inference modes when appropriately prompted. For our winning AIMO-2 submission we used an intermediate version of the 14B model and implemented various inference optimizations to accommodate the competition's time and compute constraints. We discuss model training process and evaluation results in Section~\ref{sec:model} and list Kaggle-specific optimizations in Section~\ref{sec:kaggle}.

To accelerate progress in open-source mathematical reasoning, we are releasing our code, finetuned \texttt{OpenMath-Nemotron} models, and the complete \texttt{OpenMathReasoning} dataset under a commercially permissive license.\footnote{Data and models are available at \href{https://huggingface.co/collections/nvidia/openmathreasoning-68072c0154a5099573d2e730}{https://huggingface.co/collections/nvidia/openmathreasoning-68072c0154a5099573d2e730}, our code is avaliable at \href{https://github.com/NVIDIA/NeMo-Skills}{https://github.com/NVIDIA/NeMo-Skills}}

\section{Data Preparation}
\label{sec:data}

In this section, we outline our validation and training data curation pipeline. Section~\ref{sec:problem-prep} presents our methodology for preparing a large-scale problem set for training. Section~\ref{sec:comp-math-24-25} describes our validation set collection process. Finally, Section~\ref{sec:cot-sol-prep} details our approach to synthesizing long-reasoning Chain-of-Thought (CoT) solutions.

\subsection{Problems preparation}
\label{sec:problem-prep}

We collect a large set of mathematical problems from the \href{https://artofproblemsolving.com/community}{Art of Problem Solving (AoPS) community forums}. We include all forum discussions except ``Middle School Math'', which we found to be too elementary and unhelpful for training in our preliminary experiments. After retrieving forum discussions, we implement a systematic process to extract problems and their corresponding answers. Throughout our pipeline, we utilize \texttt{Qwen2.5-32B-Instruct}~\citep{qwen2025qwen25technicalreport} for all processing steps unless specified otherwise.

\begin{enumerate}
    \item \textbf{Problem Extraction:} We prompt an LLM to identify and extract all problems from the initial forum posts (Appendix~\ref{sec:forum_problem_extraction}). While most posts contain a single problem, some include multiple problems or none at all.
    \item \textbf{Problem Classification:} Each extracted problem is classified into the following categories. We use an LLM to perform the classification:
    \begin{itemize}
        \item Proof problem or not (Appendix~\ref{sec:proof_problem_classification})
        \item Multiple choice question or not (Appendix~\ref{sec:multiple_choice_problem_classification})
        \item Binary question (yes-or-no answer) or not (Appendix~\ref{sec:binary_problem_classification})
        \item Valid problem or not (Appendix~\ref{sec:valid_problem_classification})\footnote{E.g. problems that are lacking context or referring to other problems are considered invalid.}
    \end{itemize}
    We remove all multiple-choice questions, binary questions, and invalid problems from the final dataset.
    \item \textbf{Question Transformation:} For proof questions, we convert them into answer-based questions that require similar problem-solving techniques (Appendix~\ref{sec:proof_problem_conversion}).
    \item \textbf{Answer Extraction:} For non-proof questions, we attempt to extract the final answer from the forum discussions (Appendix~\ref{sec:forum_answer_extraction})\footnote{We do not try to extract the full solution, just the final answer.}.
    \item \textbf{Benchmark Decontamination:} Following~\cite{yang2023rethinking} we use an LLM-based comparison to remove questions that closely resemble those in popular math benchmarks. 
\end{enumerate}

All prompts and scripts necessary to run the above pipeline are available in \href{https://nvidia.github.io/NeMo-Skills/openmathreasoning1/dataset/}{NeMo-Skills}. Table~\ref{tab:pipeline_stages} has a breakdown of the dataset size after each processing stage and Table~\ref{tab:dataset_comp} shows the final dataset composition. We provide a comparison with other popular datasets sourced from AoPS forums in Table~\ref{tab:aops_data_comp}.

\begin{table}[h!]
\centering
\renewcommand{\arraystretch}{1.2} 
\begin{tabular}{lc}
\hline
\textbf{Pipeline Stage} & \textbf{Data Size} \\
\hline
Original forum discussions  & 620K \\
Extracted problems & 580K \\
Removing ``bad'' problems & 550K \\
Benchmark decontamination & 540K \\
\hline
\end{tabular}
\caption{Dataset size after each processing stage.}
\label{tab:pipeline_stages}
\end{table}

\begin{table}[h!]
\centering
\renewcommand{\arraystretch}{1.2} 
\begin{tabular}{lc}
\hline
\textbf{Subset} & \textbf{Size} \\
\hline
Converted proofs  & 260K \\
With extracted answer & 190K \\
No extracted answer & \phantom{1}90K \\
\hline
Total problems & 540K \\
\hline
\end{tabular}
\caption{Final dataset composition.}
\label{tab:dataset_comp}
\end{table}

\begin{table}[h!]
\centering
\renewcommand{\arraystretch}{1.2} 
\setlength{\tabcolsep}{2pt}
\begin{tabular}{lc}
\hline
\textbf{Dataset} & \textbf{\# of Problems} \\
\hline
OpenMathReasoning (ours)  & 540K \\
AoPS-Instruct~\cite{mahdavi2025leveraging} & 650K \\
NuminaMath-1.5 (AoPS part)~\cite{numina_math_datasets} & \phantom{1}68K \\
\hline
\end{tabular}
\caption{Comparison with other datasets sourced from AoPS forums. Our work was done concurrently with~\cite{mahdavi2025leveraging} and~\cite{numina_math_datasets}.}
\label{tab:aops_data_comp}
\end{table}

\subsection{Comp-Math-24-25 Benchmark}
\label{sec:comp-math-24-25}

To create a robust validation dataset for our evaluation, we combined problems from American Invitational Mathematics Examinations (AIME) and Harvard-MIT Mathematics Tournaments (HMMT) gathered from the Art of Problem Solving forums. We restricted our selection to 2024 and 2025 competitions to minimize potential data contamination. AIME and HMMT problems were selected for our validation set due to their strong alignment with AIMO-2 competition requirements. They covered similar mathematical topics, matched the difficulty level, and were predominantly non-proof-based questions requiring single numerical answers. We excluded proof-based questions and those awarding partial credit based on estimate accuracy, as these are generally incompatible with an exact match evaluation framework. The resulting dataset, which we call \textbf{Comp-Math-24-25}, consists of 256 problems, as detailed in Table \ref{tab:validation_dataset}.

\begin{table}[h!]
\centering
\renewcommand{\arraystretch}{1.2} 
\begin{tabular}{lc}
\hline
\textbf{Problem source} & \textbf{\# of Problems} \\
\hline
AIME 2024 & \phantom{1}30 \\
AIME 2025 & \phantom{1}30 \\
HMMT Nov 2024 & \phantom{1}62 \\
HMMT Feb 2024 & \phantom{1}68 \\
HMMT Feb 2025 & \phantom{1}66 \\
\hline
\textbf{Total} & 256 \\
\hline
\end{tabular}
\caption{Composition of our Comp-Math-24-25 validation dataset.}
\label{tab:validation_dataset}
\end{table}

\subsection{Text-based Solution Synthesis}
\label{sec:cot-sol-prep}

To generate CoT solutions, we follow a common pipeline of directly prompting an existing open-weight LLM to solve problems collected in Section \ref{sec:problem-prep}. We utilize \texttt{DeepSeek-R1} and \texttt{QwQ-32B} models and generate up to 32 solution candidates for each problem in our dataset. We use temperature $0.7$, $\text{top-}p=0.95$, and limit generations to 16384 tokens. We generate more solutions for \emph{harder} problems with known answers, where the hardness was estimated by computing an average pass-rate across 32 generations from the \texttt{Qwen2.5-72B-Math-Instruct} model~\cite{yang2024qwen25mathtechnicalreportmathematical}. 

As the final filtering step we remove any solutions that do not reach the expected answer. Predicted and expected answers are compared by prompting \texttt{Qwen2.5-32B-Instruct} to judge whether they are equivalent in the context of the problem (we re-use judge prompt from~\cite{toshniwal2024openmathinstruct2}). For each problem where we weren't able to extract the final answer (and for all converted proofs) we treat the most common answer across all available solution candidates as the ground-truth. Table~\ref{tab:qwq-r1-ratio} shows the final distribution of CoT solutions in our dataset.

\begin{table}[!ht]
\centering
\renewcommand{\arraystretch}{1.2} 
\begin{tabular}{lcc}
\hline
\multirow{2}{*}{\textbf{Model}} & \multicolumn{2}{c}{\textbf{CoT solutions}} \\
 & \textbf{after filtering} & \textbf{all} \\
\hline
QwQ-32B & 0.5M & 1.0M \\
DeepSeek-R1  & 2.7M & 4.2M \\\hline
Total & 3.2M & 5.2M \\
\hline
\end{tabular}
\caption{Final distribution of CoT solutions in our dataset.}
\label{tab:qwq-r1-ratio}
\end{table}

\section{Tool-Integrated Reasoning}
\label{sec:tir}

Allowing LLMs to integrate natural language reasoning with Python code execution is a known way of improving accuracy on challenging math problems~\cite{toshniwal2024openmathinstruct, yang2024qwen25mathtechnicalreportmathematical}. However, the best open-weight reasoning models (most notably \texttt{DeepSeek-R1}~\cite{guo2025deepseek} and \texttt{QwQ-32B}~\cite{qwq32b}) are not able to directly produce such Tool-Integrated Reasoning (TIR) solutions. Our initial attempts to induce TIR generations by prompting these reasoning models with direct instructions or few-shot examples turned out to be unsuccessful. Unable to solve this via prompting, we had to develop a more elaborate pipeline for building reasoning models capable of producing TIR solutions.

In our early experiments, we noticed that when non-reasoning \textit{instruct} LLMs are trained on a limited quantity of reasoning data~\cite{ye2025limo}, they tend to retain their good instruction-following abilities. Building on this intuition, we were able to successfully prompt \texttt{LIMO-Qwen-32B}~\cite{ye2025limo} model to produce TIR solutions, but found that they tend to be \emph{low-quality} on average. The produced code was often irrelevant or was merely used to verify calculations of preceding CoT steps. To overcome this, we developed a filtering step aimed at retaining only high-quality examples where code execution provides substantial reasoning benefits. Using this filtered dataset, we then fine-tuned our reasoning model, achieving significant performance improvements over the CoT-only predecessor. Finally, we employed an iterative model improvement approach by training a more powerful TIR model in each iteration and using it to generate and filter additional TIR examples, further enhancing model performance.  In the following subsections, we detail each stage of this pipeline. 

\subsection{Instruction-following reasoning model}

Prior work~\cite{muennighoff2025s1simpletesttimescaling, ye2025limo} shows that fine-tuning on as few as 1000 samples is sufficient to make LLM produce long-CoT solutions. We hypothesize that an \emph{instruct} model fine-tuned on such a small dataset can potentially preserve its instruction-following and long-reasoning capabilities.

To test this, we prompted \texttt{LIMO-Qwen-32B} to solve the problem using Python code for the steps that require complex calculations. The zero-shot prompt we designed for this purpose is provided in Appendix \ref{sec:tir_instruction}. For roughly half of the problems, the model produced a solution that contained at least one Python code block. 
We then synthesized 1.2M solutions for \texttt{OpenMathReasoning} problems, using temperature=0.7, top-$p=0.95$, allowing maximum sequence length of 16384 tokens and stopping generations if the solution contained more than 8 code executions.

\subsection{Filtering TIR data}
\label{sec:tir-stage0}

Careful inspection of generated solutions revealed that code execution often does not benefit the solution and could easily be replaced with several simple CoT steps (see example in Appendix \ref{lst:tir-non-novel-calculation}). Instead, we want an ideal TIR solution to provide significant shortcuts by implementing otherwise infeasible brute-force approaches, e.g., using numeric solvers or conducting an exhaustive search of possible solutions. To filter unwanted code usages, we apply several filters. First, we utilize \texttt{Qwen2.5-32B-Instruct} to classify each code block by two criteria:

 \begin{itemize}
    \item  \textbf{novel calculation} / \textbf{verification}. Whether the code execution leads to a novel result or it simply verifies the previous steps (see the prompt in Appendix~\ref{sec:TIR_usage_classification}).
    \item \textbf{significant} / \textbf{moderate} / \textbf{trivial}. Whether the code implements an important part of the solution or is easily substitutable with several CoT steps (see the prompt in Appendix~\ref{sec:TIR_significance_evaluation}).
\end{itemize}

We then only keep solutions that either have at least one novel and significant code block or more than half novel and moderate code blocks. Additionally, we apply rule-based filtering and remove solutions with incorrect final answer and solutions without code execution. We also remove solutions with more than two code blocks, as we found it to be helpful in our preliminary experiments. As part of preprocessing, we also replace the tags marking the start and end of code blocks. During \textit{stage-0} generation, we instruct the model to place code between "\verb|```python|" and "\verb|```\n|", following a markdown-like style that models can easily produce; we then replace these with "<tool\_call>" and "</tool\_call>" tags, respectively, to make the code ending tag distinguishable from regular markdown and facilitate code extraction. All described filtering steps result in the TIR dataset, consisting of 15k samples, which we will refer to as \textit{stage-0 TIR data}. 

\subsection{Iterative data generation}
\label{sec:tir-self-improvement}

For the next stage of TIR solution generation, we leverage \texttt{QwQ-32B} as it proved to be a powerful yet lightweight synthetic reasoning data generator.
For this purpose, we fine-tune it on the \emph{stage-0} data for 7 epochs with a constant learning rate of 5e-6. We then synthesize solutions for \texttt{OpenMathReasoning} problems. 
We generate 700K samples and filter them down to 260K by removing incorrect solutions and solutions not using code. We find that novelty and significance filters degrade the performance at this stage, so we do not use them.

To further improve results, we repeat this process one more time using an intermediate version of our 14B model, which was finetuned on the CoT-only subset of \texttt{OpenMathReasoning} data. We train this 14B model on \texttt{QwQ-32B} solutions and then execute a final round of data generation and filtering, ultimately resulting in the final 1.7M TIR dataset.

\subsection{Controlling the number of code blocks}

We developed a simple, yet effective method to control the number of code blocks that the model can use. During all data generation stages, we format the code output as shown in Appendix \ref{sec:tir_solution_code_execution_limit}, appending additional notification warning about how many code executions are remaining. We find that model often refers to this message in its thinking process, refraining from further code usage when the limit is reached. Thus, for each problem we randomly select between 1 and 8 allowed code executions and provide this information in the prompt. We remove generations that try to use more code blocks than requested in order to reinforce the correct behavior in training. As a result, model learns to follow specified code execution limit. An example of this behavior is provided in Appendix \ref{lst:tir-code-limit-reached}.

\section{Generative Solution Selection}
\label{sec:genrm}

\begin{figure*}[t]
    \centering
    \begin{subfigure}{0.45\textwidth}
        \centering
        \includegraphics[width=\linewidth]{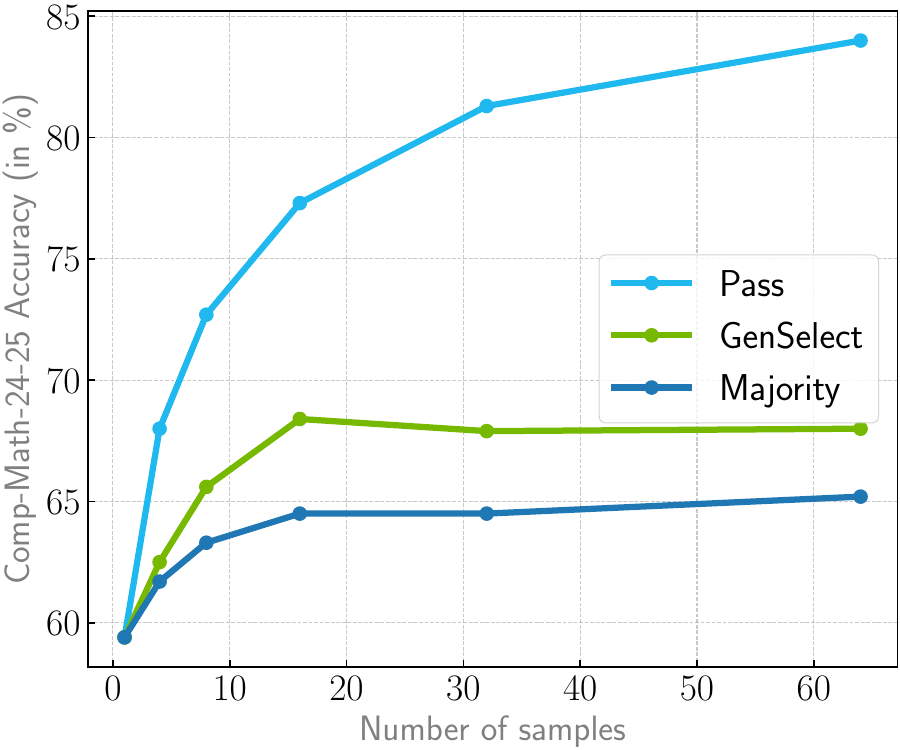}
        \caption{\centering 14B CoT}
        \label{fig_cot_genrm}
    \end{subfigure}
    \hfill
    \begin{subfigure}{0.45\textwidth}
        \centering
        \includegraphics[width=\linewidth]{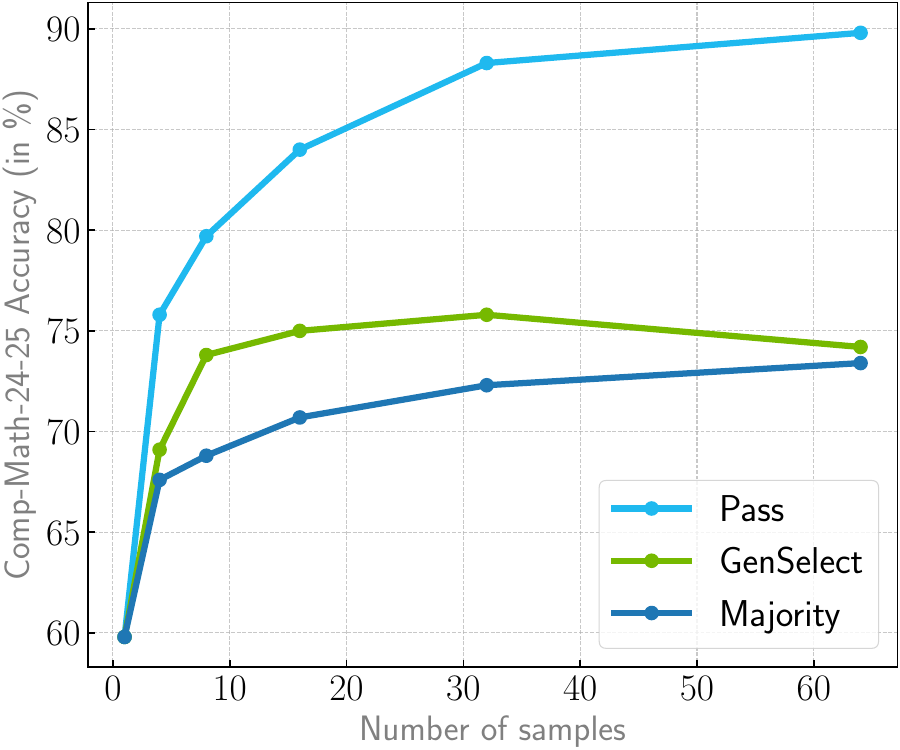}
        \caption{\centering 14B TIR}
        \label{fig_tir_genrm}
    \end{subfigure}
    \caption{Comparison of majority, GenSelect and pass metrics for different number of generation samples. To construct the input for GenSelect, we use subsets of 16 solutions (or all if fewer samples were generated). For the final answer, we perform majority@8 over the answers selected by the GenSelect. \texttt{OpenMath-Nemotron} \texttt{-14B} model is used to perform CoT, TIR, and GenSelect inference. We find that GenSelect becomes unstable when using more than 32 generations as we can no longer show all solutions in a single prompt.}
    \label{fig:genselect}
\end{figure*}

\begin{figure}[!t]
    \centering
    \resizebox{1.0\linewidth}{!}{\includegraphics{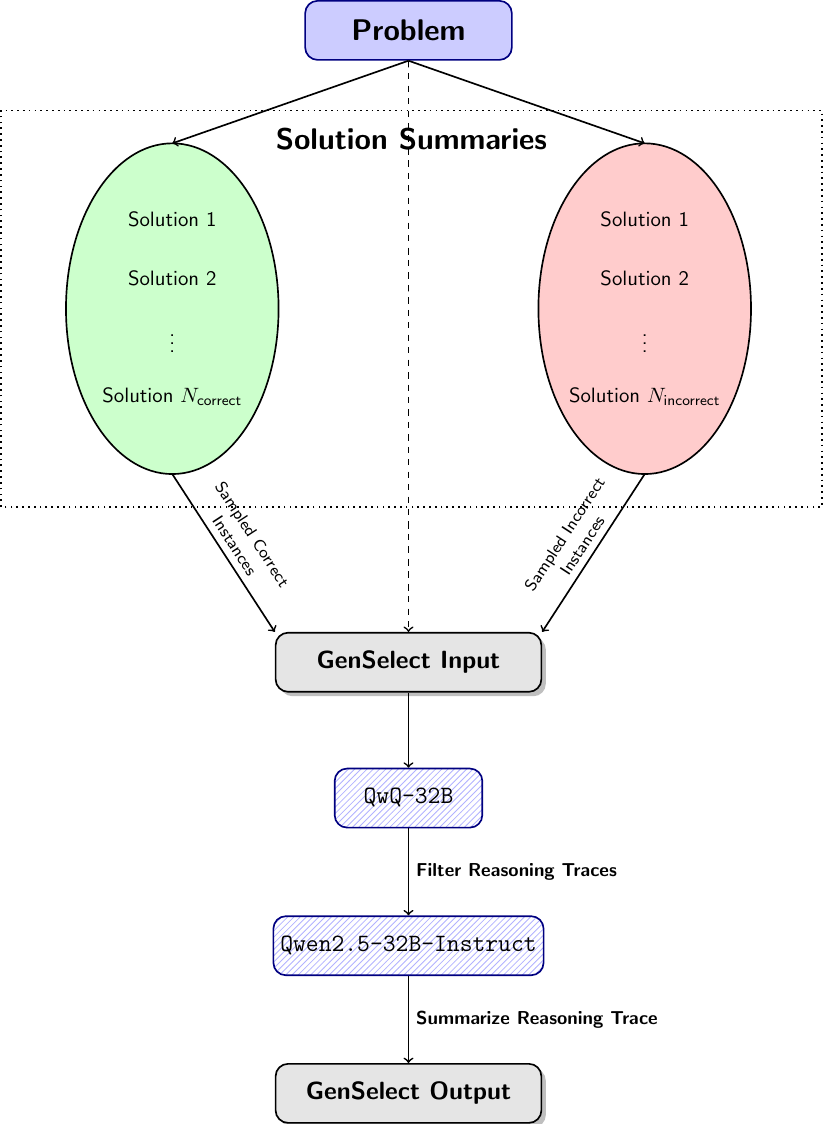}}
    \caption{Data construction pipeline of GenSelect. The GenSelect input is constructed by sampling solution summaries of both correct and incorrect instances, ensuring that the input contains at least one correct and one incorrect solution. The input is then fed to \texttt{QwQ-32B}, which is tasked with selecting the best solution among the candidate solutions. The reasoning traces that select correct solutions are summarized with \texttt{Qwen2.5-32B-Instruct}, which forms the GenSelect output.}
    \label{fig:gen_select_pipeline}
\end{figure}

We observe a considerable gap in the \texttt{majority@k} vs \texttt{pass@k} performance for our models, implying the models theoretical ability to solve far more problems than can be achieved with a majority answer. To bridge this gap, we explore training a model that, given a set of candidate solution \textit{summaries}, picks the most promising solution. In our early experiments, we found that comparing multiple solutions yields significantly better results than judging each solution in isolation. Following~\cite{zhang2024generative}, we do not change the model's architecture and instead let it reason in natural language before selecting one of the provided solutions. We detail the pipeline to prepare the training data for such \textit{selection} generations (GenSelect) in the following sections. The data construction pipeline of is shown in Figure \ref{fig:gen_select_pipeline}.




\subsection{Creating New Summaries}

\label{sec:new-summaries}
Solutions generated by reasoning models have a \emph{thinking} part and a \emph{summary} which follows it. 
We noticed that summaries generated by reasoning models, such as \texttt{DeepSeek-R1}, could be very succinct; in extreme cases, they could just be stating the final answer. Since we require a representative summary for comparing different solutions during inference, we replace the \emph{native} summary of the reasoning models by synthesizing new summaries with the \texttt{Qwen2.5-32B-Instruct} model. We synthesize four candidate summaries per solution with a maximum length of 2048 tokens. To ensure the summary is faithful, we filter out summaries where the predicted answer is different from the original solution's predicted answer. If there are no valid summaries, we discard the sample\footnote{No more than 5\% of all samples were discarded this way.}, otherwise we select the longest summary to replace the original summary. We regenerate summaries for the entire OpenMathReasoning dataset using this process, so that models trained on it can produce these summaries directly. See Appendix~\ref{sec:summary_comparison} for a comparison between one-word \texttt{DeepSeek-R1} summary and a new one generated by \texttt{Qwen2.5-32B-Instruct}.

\subsection{Generating Selection Candidates}

We discover that modest accuracy gains over majority voting can be achieved by simply presenting new solution summaries to reasoning models and prompting them to compare and select one (see prompt in Appendix ~\ref{sec:math_genrm_selection}). Building on this observation, we develop the following pipeline to generate training data for this GenSelect inference.

For each problem in the \texttt{OpenMathReasoning} dataset, we randomly sample between 2 and 16 candidate solution summaries. We ensure that each sample group contains at least one correct and one incorrect solution. This process is repeated until we obtain 8 distinct comparison groups for each problem. Using the GenSelect prompt (Appendix~\ref{sec:math_genrm_selection}), we then task \texttt{QwQ-32B} with selecting the most promising solution from each group. This procedure generates 1M selections, which we subsequently filter down to 565K by eliminating any instances where incorrect solutions were chosen.


\subsection{Reducing computational cost}

While this dataset is suitable for training, the comparison generations can be as long as the original solutions, making GenSelect inference computationally expensive. To address this challenge, we explored training models to directly generate the final comparison \textit{summary} rather than learning the full reasoning trace. Consistent with our previous observations, the natural comparison summaries produced by \texttt{QwQ-32B} proved suboptimal. We therefore again used \texttt{Qwen2.5-32B-Instruct} to regenerate all comparison summaries (see the prompt in Appendix~\ref{sec:gen_summarization}) and trained our models using these summarized comparisons. Our early experiments revealed only a small reduction in accuracy (1--2\%) compared to models trained on the whole reasoning traces.
This final setup makes GenSelect inference remarkably efficient compared to the original long-reasoning generations. With output tokens capped at 2048, most computation occurs in a highly-parallelizable pre-filling phase. Since each solution summary is similarly limited to 2048 tokens, the total input context cannot exceed 32768 tokens when using the maximum of 16 solutions per problem. Although more than 16 solutions could theoretically be included in a prompt, we generally observe diminishing returns as the context becomes too large. For scenarios requiring evaluation of more solution candidates, we propose sampling 16 solutions multiple times and then performing majority voting to determine the final answer. Nevertheless, our findings indicate that the most significant accuracy improvements occur when GenSelect is applied to a smaller number of generations (Figure~\ref{fig:genselect}).

\section{OpenMath-Nemotron models}
\label{sec:model}

In this section we present the training and evaluation details of our \texttt{OpenMath-Nemotron} series of models.

\begin{figure*}[t]
    \centering
    \begin{subfigure}{0.45\textwidth}
        \centering
        \includegraphics[width=\linewidth]{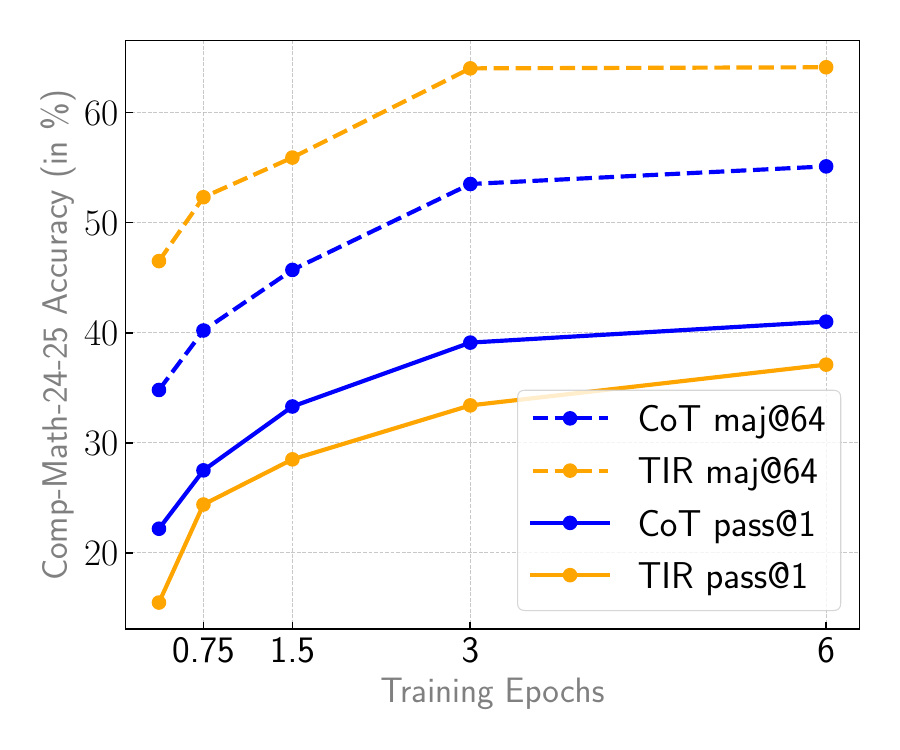}
        \caption{\centering 1.5B}
        \label{fig:1a}
    \end{subfigure}
    \hfill
    \begin{subfigure}{0.45\textwidth}
        \centering
        \includegraphics[width=\linewidth]{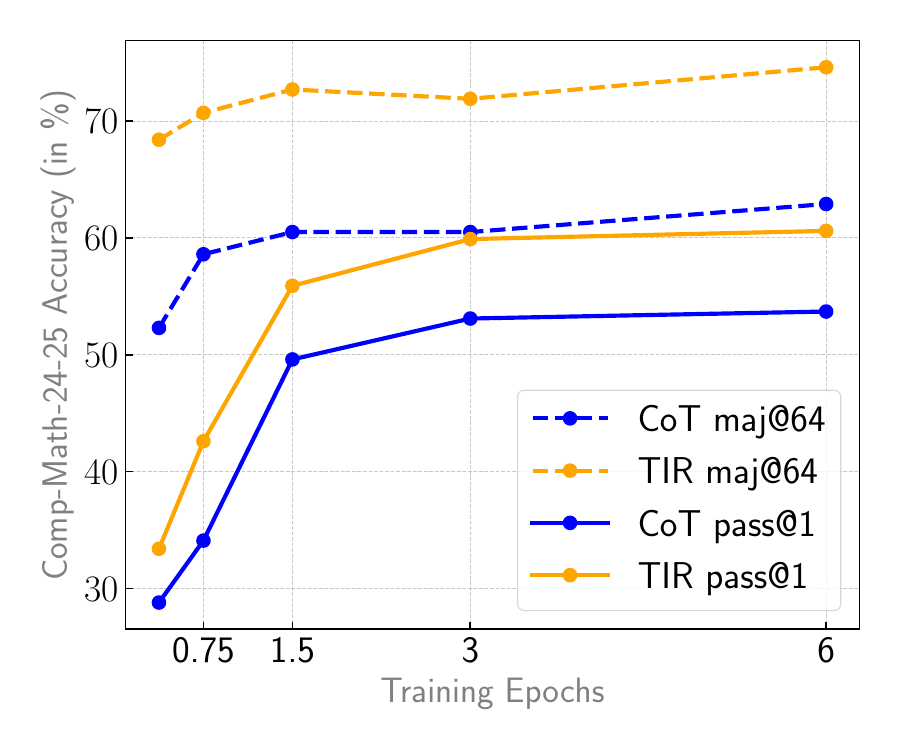}
        \caption{\centering 14B}
        \label{fig:1b}
    \end{subfigure}
    \caption{Accuracy improvement through the course of training. We observe that smaller models need to be trained for longer to achieve meaningful improvements.}
    \label{fig:training}
\end{figure*}

\subsection{Training}

To build our final models we perform supervised-finetuning (SFT) on a series of \texttt{Qwen2.5-Base} models (1.5B, 7B, 14B and 32B)~\cite{qwen2025qwen25technicalreport}. For 1.5B and 7B models, we start from the special model versions finetuned for mathematical reasoning tasks~\cite{yang2024qwen25mathtechnicalreportmathematical}. Unlike general Qwen2.5 models, the math versions only support a limited context window of 4096 tokens, which is inadequate for the long-reasoning generations. To overcome this, we follow~\cite{bloc97_ntkaware_2023} and change RoPE~\cite{su2021roformer} base to 500K.

All models are trained for six epochs on a combination of three tasks: CoT solution generation, TIR solution generation, and GenSelect, where the task is to select one correct solution out of multiple candidates. Each task is defined by a unique prompt that we can use at inference time to switch between different generation modes (see prompts in Appendix~\ref{sec:inference-prompts}). We found that training on a mix of all tasks results in a similar accuracy compared to training on each task sequentially (first CoT, then TIR, then GenSelect). The total SFT dataset size is 5.5M samples (3.2M CoT, 1.7M TIR, and 566K GenSelect).

We train all models using AdamW optimizer~\cite{loshchilov2017decoupled} with weight decay of $0.01$ and a cosine learning rate decay schedule with a 10\% linear warmup. We use a starting learning rate of 3e-4 for 1.5B, 2e-4 for 7B and 1e-4 for 14B and 32B models. The final learning rate is set to be 1000 times smaller. We use batch size of 1024 samples and leverage sequence packing and context parallelization techniques from NeMo-Aligner~\cite{shen2024nemo} that significantly accelerate training on the long-reasoning data. Following~\cite{toshniwal2024openmathinstruct2} we save 4 equally spaced checkpoints during the training runs, which are averaged to create the final model. We show the accuracy on the Comp-Math-24-25 benchmark (Section~\ref{sec:comp-math-24-25}) of intermediate 1.5B and 14B model checkpoints in Figure~\ref{fig:training}.

After the first round of training, we perform another SFT on a subset of harder problems. These problems are selected only from forums discussing Olympiad math and we discard any problems for which \texttt{Qwen2.5-Math-72B-Instruct} TIR model has a pass-rate bigger than 0.3 out of 32 generations. Additionally, we filter any solutions that have fewer than 5000 tokens. The total SFT data size of this harder set is 2.2M samples. We follow the same setup as for the first round of SFT except we train for 4 epochs instead of 6. We do this second round of training for all models except 32B as we found some degradation in results. Models' accuracy after the first and second round of training is presented in Table~\ref{tab:training-accuracy}. We find that CoT results tend to significantly improve while TIR results stay stable or slightly degrade.

\begin{table}[h!]
\centering
\renewcommand{\arraystretch}{1.2} 
\begin{tabular}{rcc}
\hline
\textbf{Model} & \textbf{First SFT} & \textbf{Second SFT} \\
\hline
1.5B CoT & 55.1 & 58.2 \\
1.5B TIR & 64.1 & 64.5 \\
7B CoT & 61.3 & 62.5 \\
7B TIR & 71.1 & 70.7 \\
14B CoT & 62.9 & 65.2 \\
14B TIR & 74.6 & 73.4 \\
\hline
\end{tabular}
\caption{Accuracy with majority@64 on the Comp-Math-24-25 benchmark after the first and second SFT rounds. We see significant gains for CoT generations and comparable results for TIR generations.}
\label{tab:training-accuracy}
\end{table}

\subsection{Results}

\begin{table*}[t]
\centering
\renewcommand{\arraystretch}{1.2} 
\begin{tabular}{lcccc}
\hline
\multirow{2}{*}{\textbf{Model}} & \multicolumn{3}{c}{\textbf{Comp-Math-24-25}} & \multirow{2}{*}{\textbf{HLE-Math}} \\
 & \textbf{AIME24} & \textbf{AIME25} & \textbf{HMMT-24-25} &  \\
 \hline
DeepSeek-R1-Distill-Qwen-1.5B & 26.8 (60.0) & 21.4 (36.7) & 14.2 (26.5) & 2.9 (5.0) \\
OpenMath-Nemotron-1.5B CoT & 61.6 (80.0) & 49.5 (66.7) & 39.9 (53.6) & 5.4 (5.4) \\
OpenMath-Nemotron-1.5B TIR & 52.0 (83.3) & 39.7 (70.0) & 37.2 (60.7) & 2.5 (6.2) \\
\multicolumn{1}{c}{+ Self GenSelect} & 83.3 & 70.0 & 62.2 & 7.9 \\
\multicolumn{1}{c}{+ 32B GenSelect} & 83.3 & 70.0 & 62.8 & 8.3 \\
\hline
DeepSeek-R1-Distill-Qwen-7B & 54.4 (80.0) & 38.6 (53.3) & 30.6 (42.9) & 3.3 (5.2) \\
OpenMath-Nemotron-7B CoT & 74.8 (80.0) & 61.2 (76.7) & 49.7 (57.7) & 6.6 (6.6) \\
OpenMath-Nemotron-7B TIR & 72.9 (83.3) & 57.5 (76.7) & 54.6 (66.3) & 7.8 (10.8) \\
\multicolumn{1}{c}{+ Self GenSelect} & 86.7 & 76.7 & 68.4 & 11.5 \\
\multicolumn{1}{c}{+ 32B GenSelect} & 86.7 & 76.7 & 69.9 & 11.9 \\
\hline
DeepSeek-R1-Distill-Qwen-14B & 65.8 (80.0) & 48.4 (60.0) & 40.1 (52.0) & 4.2 (4.8) \\
OpenMath-Nemotron-14B-MIX (kaggle) & 73.7 (86.7) & 57.9 (73.3) & 50.5 (64.8) & 5.7 (6.5) \\
OpenMath-Nemotron-14B CoT & 76.3 (83.3) & 63.0 (76.7) & 52.1 (60.7) & 7.5 (7.6) \\
OpenMath-Nemotron-14B TIR & 76.3 (86.7) & 61.3 (76.7) & 58.6 (70.9) & 9.5 (11.5) \\
\multicolumn{1}{c}{+ Self GenSelect} & 86.7 & 76.7 & 72.4 & 14.1 \\
\multicolumn{1}{c}{+ 32B GenSelect} & 90.0 & 76.7 & 71.9 & 13.7 \\
\hline
QwQ-32B & 78.1 (86.7) & 66.5 (76.7) & 55.9 (63.3) & 9.0 (9.5) \\
DeepSeek-R1-Distill-Qwen-32B & 66.9 (83.3) & 51.8 (73.3) & 39.9 (51.0) & 4.8 (6.0) \\
OpenMath-Nemotron-32B CoT & 76.5 (86.7) & 62.5 (73.3) & 53.0 (59.2) & 8.3 (8.3) \\
OpenMath-Nemotron-32B TIR & 78.4 (93.3) & 64.2 (76.7) & 59.7 (70.9) & 9.2 (12.5) \\
\multicolumn{1}{c}{+ Self GenSelect} & 93.3 & 80.0 & 73.5 & 15.7 \\
\hline
DeepSeek-R1 & 79.1 (86.7) & 64.3 (73.3) & 53.0 (59.2) & 10.5 (11.4) \\
\hline
\end{tabular}
\caption{Evaluation results on mathematical benchmarks. All models are evaluated with a maximum of 32768 output tokens, temperature of $0.6$, and top-p $0.95$. We present metrics as pass@1 (maj@64) where pass@1 is an average accuracy across 64 generations and maj@64 is the result of majority voting. The 14B model used in our kaggle submission is denoted as (kaggle). For HMMT and HLE-Math benchmarks we use LLM-judge setup of~\cite{toshniwal2024openmathinstruct2} to verify the answers. To construct the input for GenSelect, we use subsets of 16 solutions from the set of 64 solutions. We repeat this 64 times and perform majority voting over the answers selected by the GenSelect. }
\label{tab:eval-results}
\end{table*}

Final evaluation results of our models are presented in Table~\ref{tab:eval-results}. In addition to Comp-Math-24-25, introduced in Section \ref{sec:comp-math-24-25}, we use Humanity's Last Exam dataset \cite{phan2025humanitysexam}. We only evaluate on a subset consisting of 975 text-only problems from ``Math'' category. We refer to it as HLE-Math.

We notice that despite being superior in majority@$k$ setting with TIR prompt, smaller models perform on par or even worse in pass@1, compared to CoT prompt. The results in Table \ref{tab:unfinished-solutions} suggest that the reason is that with the TIR prompt there are more unfinished solutions across all model sizes, with 1.5B clearly standing out. We hypothesize that the reason behind this is that smaller models are less consistent in using tools effectively.

\section{Kaggle submission}
\label{sec:kaggle}

In this section, we present the details of our winning submission to the AI Mathematical Olympiad - Progress Prize 2 (AIMO-2) \cite{frieder2024aimo}. AIMO-2 is a competition organized by the AIMO Prize team and hosted on Kaggle \footnote{\href{https://www.kaggle.com/competitions/ai-mathematical-olympiad-progress-prize-2}{https://www.kaggle.com/competitions/ai-mathematical-olympiad-progress-prize-2}}. The competition featured 110 challenging math problems: 10 were made publicly available as a reference subset, while the remaining problems were split between the public and private leaderboards. Submitted solutions were evaluated under strict computational constraints: a 5-hour time limit in an offline Jupyter notebook environment powered by four L4 GPUs. Our 1st-place submission correctly solved 34 out of 50 questions on the private leaderboard.

\subsection{Training recipe}

For our winning Kaggle submission we used a somewhat different training recipe that we detail in this section. We first trained \texttt{Qwen2.5-14B-Base} model for 8 epochs on a 2.2M subset of CoT solutions, excluding any converted proof problems. We only used \texttt{DeepSeek-R1} solutions for this training.
This is followed by a light-weight fine-tuning on 15k \textit{stage-0 TIR} samples. The process for collecting these samples is detailed in section~\ref{sec:tir-stage0}. We train TIR model for 400 steps with a constant learning rate of 1e-5 and use the last checkpoint without averaging.
We then merge CoT and TIR checkpoints as it both improves accuracy and speeds up generation by reducing solution length and number of code executions. We did not use GenSelect training or inference for the Kaggle submission.

\subsection{Model Merging}

In this competition, we explored various methods for merging two LLMs with CoT and TIR behaviors. Our primary goal was to effectively combine the distinct strengths of these two fine-tuning stages to enhance model performance. We experimented with several merging techniques from mergekit~\cite{goddard2024mergekit} package. Surprisingly, the most effective approach turned out to be a simple linear combination of the two checkpoints: the CoT checkpoint used before TIR fine-tuning and the best TIR checkpoint attained thereafter. This strategy allowed us to control the extent to which each stage influenced the final model's behavior. Table~\ref{tab:model-merging} provides the accuracy, as well as the generation length and code usage statistics of the models before and after the described merging procedure.



\subsection{Inference Optimizations}

The strict time limits of the competition presented a serious constraint. An extra requirement was that problems had to be answered one-at-a-time making it harder to parallelize computation and allocate time. To overcome these challenges we implemented several optimizations that maximize inference efficiency while maintaining output quality.

\subsubsection{TensorRT-LLM Optimization}
Pretrained models were converted to TensorRT engines using TensorRT-LLM \cite{tensorrtllm}. TensorRT-LLM's in-flight batching boosts throughput by dynamically grouping inference requests, releasing each sample as soon as it completes—reducing latency and optimizing GPU utilization. Since samples are processed independently, batches can mix different prompts or inference parameters seamlessly. TensorRT-LLM includes a number of other optimizations such as custom attention kernels and paged KV caching.

Quantization involves a speed-accuracy tradeoff, as outlined in TensorRT-LLM's best practices \cite{tensorrtbestpractices}. We prioritized int8 weight-only (W8A16) and FP8 quantization, which delivered faster inference than BF16 with minimal accuracy loss. The reduced weight size also freed up memory for larger key-value caches.

\subsubsection{Speculative Decoding}

To accelerate inference, we employ ReDrafter \cite{redrafter}, a recurrent speculative decoding technique that uses an RNN-based drafter to propose and verify multiple tokens per decoding step. We trained a drafter capable of proposing up to three tokens at each step, with all three tokens being accepted in approximately 65\% of the steps. 

For training ReDrafter, we sampled a random subset of 100k problems from the \texttt{OpenMathReasoning} dataset. With the target model, we generated one solution per problem, leveraging the resulting data to train the drafter. 

Table~\ref{tab:quant-perf-metrics} presents an evaluation of various quantization techniques and the speculative decoding method, analyzing their impact on both the inference speed and the accuracy.



We experimented with various sampling parameters but observed minimal differences in the results. We thus based our winning submission on an \textit{almost greedy} search strategy by setting temperature to 0 and enabling the \texttt{redrafter\_greedy\_search} parameter. Despite these settings TensorRT-LLM still produced varying outputs within a single batch of identical prompts. We did not investigate this behavior in detail, but we suspect that it could be related to an accumulation of small numerical errors which cause a few tokens to be different early on in the generation. This difference then accumulates over many tokens resulting in a substantially diverse solution set at the end. Ultimately, we chose this approach because it provided more stable results at small batch sizes and offered a small improvement in the speed of speculative decoding.

\subsubsection{Model Serving}
Models were served via a FastAPI backend powered by Nemo-Skills \cite{nemoskills}, which supports time-constrained generation. This allowed us to dynamically limit response times per question—if an answer wasn't completed within the window, we returned early to check for extractable results.

Nemo-Skills' async generation enabled batched processing with early stopping. For example, in a batch of 16, if the first 4-5 completions agreed on the final answer, we canceled the remaining generations and proceeded to the next question. We also mitigated stragglers---samples that ran significantly longer than others---by canceling the last $n$ pending requests once the rest finished. This early stopping increased response relevance as shorter answers tended to be higher quality.

\begin{table}[t]
\begin{tabular}{rcc}
\hline
\textbf{Model} & \textbf{Prompt} &  \textbf{Unfinished (in \%)} \\
\hline
 1.5B &    &  \phantom{1}2.23 \\
   7B &    CoT &  \phantom{1}0.98 \\
  14B &    &  \phantom{1}1.13 \\
\hline
 1.5B &    & 40.31 \\
   7B &    TIR &  \phantom{1}6.45 \\
  14B &    &   \phantom{1}4.06 \\
\hline
\end{tabular}
\caption{Percentage of unfinished solutions on the Comp-Math-24-25 dataset. We generate 32k tokens and consider solution unfinished if it does not contain ``$\backslash$boxed''.}
\label{tab:unfinished-solutions}
\end{table}

\begin{table}[t]
\centering
\renewcommand{\arraystretch}{1.2} 
\small
\setlength{\tabcolsep}{2pt}
\begin{tabular}{p{2cm}ccc}
\hline
\textbf{Method} & \textbf{Speed (tok/s)} & \textbf{AIME24} & \textbf{AIME25} \\
\hline
BF16 & 210 & 82.7 & 66.7 \\
W8A16 (int8) & 315 & 82.7 & 66.7 \\ 
W4A16 (int4) & 436 & 72.7 & 60.7 \\
FP8 & 310 & 83.3 & 68.7 \\
\textbf{FP8$+$ReDrafter} & 554 & 81.3 & 71.3  \\
\hline
\end{tabular}
\caption{Submission pipeline with different optimizations methods benchmarked on 4 x L4 GPU. Reported scores are maj@12 on the merged model averaged over 5 runs each. \textbf{Bold} indicates configuration used in our winning submission.}
\label{tab:quant-perf-metrics}
\end{table}

\begin{table}[t]
\centering
\small
\setlength{\tabcolsep}{2pt}
\renewcommand{\arraystretch}{1.2} 
\begin{tabular}{p{2.9cm}cccc}
\hline
\textbf{Model} & \textbf{maj@16} & \textbf{pass@16} & \textbf{length} & \textbf{code} \\
\hline
CoT & 62.9 & 76.2 & 11203 & - \\
TIR & 66.8 & 80.1 & 15834 & 2.73 \\
CoT*0.3 + TIR*0.7 & 69.1 & 81.3 & 12489 & 0.85 \\
\hline
\end{tabular}
\caption{Accuracy and generation statistics of merged models on Comp-Math-24-25 dataset. \textbf{length} indicates the average number of tokens per solution, while \textbf{code} refers to the average number of code executions per solution.}
\label{tab:model-merging}
\end{table}

\subsubsection{Time Management}
\label{sec:time-management}
A buffering strategy was implemented, allocating 350 seconds per question as the base time limit. If a question completed early, the unused time was added to a shared buffer. The next question could then draw up to 210 extra seconds from this buffer, allowing a maximum of 560 seconds when combined with its base allocation.

\subsubsection{Code Execution Integration}
For tool-calling capabilities, we used Nemo-Skills's code execution wrapper to enable tool-integrated reasoning. A Flask-based sandbox environment handles parallelized Python execution for each inference thread, processing LLM-generated code blocks with strict safeguards:
\begin{itemize}
    \item Maximum 6 code calls per generation cycle
    \item 2 second timeout for each code execution
    \item Only the first 200 characters of the code output were shown back to the LLM
\end{itemize}
The system feeds back either execution results or error traces into the generation process, enabling iterative reasoning while maintaining computational efficiency.

\subsection{Discussion}

Our Kaggle submission is based on an early development version of the final \texttt{OpenMath-Nemotron-14B} model. This model was trained on a smaller dataset, did not have GenSelect capability, and could not switch between CoT and TIR modes by changing the prompt. While we did have a much better checkpoint towards the end of the competition, we were ultimately unable to make a high-scoring submission with it. In this section, we explore several potential explanations for why this happened.

\textbf{High variance in scores.} The competition rules allow only a single submission per day. Since the public leaderboard consists of only 50 problems presented in random order, we observed substantial variance across our submissions. This made it hard to make quick decisions on which directions to prioritize, especially in cases when our local evaluations disagreed with the leaderboard scores.

\textbf{Focus on smaller models.} As shown in Table~\ref{tab:eval-results} \texttt{OpenMath-Nemotron-7B} model performs comparably or better than the 14B model used in Kaggle. Observing this towards the end of the competition, we tried to prioritize submissions with the smaller model, allowing it more generations, and also increased the maximum generation length. Yet we were unable to see comparable leaderboard scores. This discrepancy suggests that either our local evaluation set differs substantially from what was used in Kaggle, or that the smaller models struggle with a few particularly challenging problems---a limitation difficult to detect through aggregate benchmark scores alone.

\textbf{Longer average generations.} Our local evaluations always had a fixed token budget for each generation. However, the time management logic implemented in Kaggle (Section~\ref{sec:time-management}) heavily relied on solving easy problems quickly to allocate more time for challenging ones. Interestingly, we discovered that although our final models achieved superior scores within the same token budget, they produced around $10\%$ more tokens \textit{on average}. Not realizing this early enough, we were unable to fix this undesirable feature before the end of the competition.
\section{Related Work}
\subsection{Tool Integration Reasoning}
Tool-augmented approaches to mathematical problem solving have advanced rapidly in recent years. A seminal contribution by Chen et al. \citep{DBLP:journals/tmlr/ChenM0C23} introduced the Program of Thoughts (PoT) framework, which integrates natural language with executable code to support step-by-step reasoning through a hybrid of textual and programmatic logic. Building on this foundation, subsequent research has focused on developing both datasets and models that facilitate tool-integrated reasoning.

On the data side, OpenMathInstruct-1 \citep{toshniwal2024openmathinstruct} offers 1.8 million instruction-tuning examples derived from code interpreters across benchmarks such as GSM8K and MATH. Similarly, InfinityMATH \citep{zhang2024infinitymath} introduces 100K instances of programmatic reasoning, while MARIO \citep{liao2024mario} combines model reasoning with tool outputs, accompanied by a dataset constructed from GSM8K \cite{cobbe2021gsm8k} and MATH \cite{hendrycks2021measuringmathematicalproblemsolving}. These resources have significantly enriched the training landscape for tool-augmented reasoning systems.

On the modeling side, Qwen2.5 \cite{yang2024qwen25mathtechnicalreportmathematical} introduced a series of models with strong mathematical reasoning capabilities, supporting advanced techniques like Chain-of-Thought (CoT) and Tool-Integrated Reasoning (TIR). Gao et al.\ \cite{gao2024efficient} proposed a two-stage method: training large language models to generate reasoning chains, and then invoking domain-specific tools to execute each step by injecting the necessary knowledge. Xiong et al. \cite{xiong2024building} proposed a multi-turn, online, iterative direct preference learning framework tailored to this unique context. By incorporating feedback from code interpreters during the training process, their approach achieves significant performance improvements on the MATH benchmark. Wu et al. \cite{wu2025agentic} dynamically integrate web search, code execution, and structured reasoning with contextual memory to tackle complex problems that demand deep research and multistep logical deduction. Li et al.\ \cite{li2025torl} developed a Tool-Integrated Reinforcement Learning framework that autonomously utilizes computational tools by scaling reinforcement learning directly from base models, and demonstrate substantial improvements compared to RL without tools. 

\subsection{Generative Reward Models}
Conventional reward models and verifiers are often trained as discriminative binary classifiers, underutilizing the generative strengths of large language models (LLMs). To address this, Generative Reward Models (GenRM) \cite{mahangenerative}, introduced by Mahan et al., reformulate verification as a generation task—using the log probabilities of tokens like “Yes” or “No” to represent correctness. This framing allows GenRM to better exploit LLMs’ natural language generation capabilities, leading to improved alignment with human judgments across both in-distribution and out-of-distribution tasks. 
Concurrently, Zhang et al. \cite{zhang2024generative} introduced Generative Verifiers, training CoT-GenRM with a supervised fine-tuning (SFT) objective to serve as a verifier for mathematical reasoning. Building on a similar motivation, Ankner et al. \cite{ankner2024critique} combined Chain-of-Thought (CoT) reasoning generation with Bradley-Terry reward modeling, enabling reward models to explicitly reason about response quality before assigning scores. Extending this line of work, Wang et al. \cite{wang2024self} proposed self-taught evaluators, jointly training generative models and LLM-as-a-Judge frameworks to produce both intermediate reasoning traces and final judgments. In related approaches, Wang et al. \cite{wang2024direct} trained large language models as generative judges by leveraging Direct Preference Optimization (DPO) on both positive and negative data, demonstrating improved evaluation performance across diverse tasks. Wu et al. \cite{wu2024meta} introduced a Meta-Rewarding step in the self-improvement process, enabling the model to evaluate its own judgments and use the feedback to refine its evaluation capabilities.

\subsection{Math Reasoning Datasets}
In the pursuit of improving mathematical reasoning in large language models, researchers have recently introduced several large-scale, high-quality datasets. Skywork-MathQA \citep{zeng2024skyworkmathdatascalinglaws} stands out with its 2.5 million question-answer pairs, generated using a trio of augmentation methods and built upon a varied set of foundational problems. Complementing this, NuminaMath \citep{li2024numinamath} offers 860K challenging competition-style math problems, each carefully annotated with step-by-step reasoning chains \citep{DBLP:conf/iclr/0002WSLCNCZ23}, enabling more interpretable and structured model outputs. 

More recent work has focused on advancing the complexity and depth of reasoning. New datasets have emerged that emphasize challenging questions paired with rich, multi-step reasoning traces, pushing models to handle more sophisticated mathematical thought processes. BackMATH  was introduced in \cite{zhang2025backmath}, a novel dataset centered on backward reasoning. It contains approximately 14K problems specifically designed to support backward problem-solving, along with 100K detailed reasoning steps that trace the reverse logical flow from solution to problem. The OpenR1 team released OpenR1-Math-220K \cite{openr1_math_2025}, a large-scale dataset for mathematical reasoning comprising 220K math problems. Each problem includes two to four reasoning traces generated by DeepSeek R1, based on problems from NuminaMath 1.5 \cite{numina_math_datasets}. In addition, Zhao et al. \cite{zhao20251} presented AM-DeepSeek-R1-Distilled, a large-scale dataset featuring 1.4 million question-response pairs with associated thinking traces for general reasoning tasks. This dataset is composed of high-quality, challenging problems aimed at advancing reasoning capabilities. Following a similar direction, Liu et al. \cite{Chinese-Data-Distill-From-R1} introduced a Chinese version of the DeepSeek-R1 distilled dataset, consisting of 110K question-solution pairs. The DolphinR1 team \cite{dolphinr1_2025} released a dataset of 800K samples, combining outputs from various reasoning models, including DeepSeek-R1, Gemini 2.0 Flash Thinking, and Dolphin Chat.

\section{Conclusion}

In this paper, we present our winning submission to the AIMO-2 competition and a pipeline for developing state-of-the-art mathematical reasoning models. Our contributions can be summarized as follows:

\begin{itemize}
    \item We develop a method to combine code execution with long chain-of-thought (CoT) generations to produce tool-integrated reasoning (TIR) solutions.
    \item We create a pipeline for training models to generate samples that select the most promising solution from multiple candidates (GenSelect).
    \item We release a large-scale \texttt{OpenMathReasoning} dataset. It contains 540K unique mathematical problems, 3.2M long chain-of-thought (CoT) solutions, 1.7M long tool-integrated reasoning (TIR) solutions, and 566K generative solution selection (GenSelect) traces.
    \item We release a series of \texttt{OpenMath-Nemotron} models capable of operating in CoT, TIR, or GenSelect inference modes. With this release, we establish a new state-of-the-art in mathematical reasoning among open-weight models.
\end{itemize}

\bibliographystyle{plain}  
\bibliography{paper}  

\appendix

\onecolumn
\section{Problem Preparation Prompts}
\label{sec:LLM_prompts}

\subsection{Binary Problem Classification}
\label{sec:binary_problem_classification}

\begin{tcolorbox}[breakable,width=\textwidth,colback=white,colframe=NvidiaGreen,title={\centering \large  \textbf{Prompt: Binary Problem Classification}}]
\footnotesize                  
\begin{lstlisting}[
breaklines=true, postbreak={},breakindent=0pt, 
label={lst:math-prompt}]
I will provide a math problem, and you need to determine whether it is a binary question.
Respond only with 'binary' if the problem meets the criteria, and 'not binary' otherwise.

A problem qualifies as a binary question if and only if:

1. The problem explicitly asks for a binary response, such as "yes or no", "true or false", or another equivalent two-choice response.
2. The problem is phrased as a question or statement that naturally leads to a binary response (e.g., "Is this true?" or "Determine whether the statement is true or false").

If the problem does not explicitly ask for a binary response, even if it can be interpreted that way, it should be classified as 'not binary question'.


Here are a few examples.

Example 1

Problem:
Is it true that $0.4395308999999\\ldots = 0.4395309$?

Output: binary



Example 2

Problem:
Write first several terms of a geometric progression in which the difference between the third and first terms is equal to 9, and that between the fifth and third terms equal 36.

Output: not binary



Example 3

Problem:
Solve the following equations: $\frac{{\sin(60^\circ+x)+\sin(60^\circ-x)}}{{2}} =\frac{{\tan x}}{{(1+\tan^2 x)^2}}+\frac{{\cot x}}{{(1+\cot^2 x)^2}}$

Output: not binary




Example 4

Problem:
Given the quadratic expression \\( ax^2 + bx + c \\) with coefficients \\( a, b, c \\) such that \\( b - c > a \\) and \\( a \\neq 0 \\), is it true that the equation \\( ax^2 + bx + c = 0 \\) always has two distinct real roots?

Output: binary



Example 5:
Problem:
Can the vertices of a cube be colored in red, yellow, and blue such that every set of four coplanar vertices contains all three colors?

Output: binary



Example 6:
Problem:
Can the numbers \\(\\frac{{14x + 5}}{{9}}\\) and \\(\\frac{{17x - 4}}{{12}}\\) both be integers for some integer \\(x\\)? If so, find that integer.


Output: not binary


Example 7:
Problem:
Can the distances from a point on the plane to the vertices of a certain square be equal to $1, 1, 2,$ and $3$?


Output: binary




Now here is the problem you need to extract the answer from.


Problem:
{problem}

Output:
\end{lstlisting}
\end{tcolorbox}

\subsection{Valid Problem Classification}
\label{sec:valid_problem_classification}

\begin{tcolorbox}[breakable,width=\textwidth,colback=white,colframe=NvidiaGreen,title={\centering \large  \textbf{Prompt: Valid Problem Classification}}]
\footnotesize                  
\begin{lstlisting}[
breaklines=true, postbreak={},breakindent=0pt, 
label={lst:math-prompt-invalid}]
I will provide a problem statement from a math forum. Your task is to determine whether it is a valid, solvable math problem based on the given text.

Respond with 'not invalid' if the problem meets all of the following conditions:

1. It is a well-defined math question, such as solving an equation, finding a minimum, computing an expression, or proving a result.
2. It contains enough information to be solved using standard mathematical techniques, even if the solution requires advanced concepts (e.g., limits, logarithms, recursion).
3. It is not just a conceptual or definitional question (e.g., "What does the notation  mean?" is not a valid math problem).
4. It does not rely on external resources such as images or missing context.

Otherwise, respond with 'invalid', but only if there is a clear and strong reason why the problem cannot be solved. If you are uncertain, default to 'not invalid'.

Important Notes:

1. The vast majority (>99%) of problems will be valid math problems.

2. Only extremely rare cases are invalid, such as: Problems relying on external images or missing definitions. Vague or incomplete statements that cannot be interpreted mathematically. Open-ended conceptual discussions rather than problem-solving.

3. A problem is still valid even if solving it requires advanced methods like recursion, limits, or logarithms.

4. Do not evaluate whether the problem has a solution or not.

5. Do not analyze the difficulty of the problem or the methods required to solve it.

6. Only check whether it is a well-formed math problem that can be meaningfully interpreted.



Here are a few examples.

Example 1

Problem:
Solve the equation \\(\\log(x - 2)(2x - 3) = \\log(x^2)\\).

Output: not invalid



Example 2

Problem:
Solve the math problem found on Facebook (image provided)

Output: invalid



Example 3

Problem:
Solve the following equations: $\frac{{\sin(60^\circ+x)+\sin(60^\circ-x)}}{{2}} =\frac{{\tan x}}{{(1+\tan^2 x)^2}}+\frac{{\cot x}}{{(1+\cot^2 x)^2}}$

Output: not invalid



Example 4

Problem:
Find the area of square T?

Output: invalid



Example 5:

Problem:
Provide another example of a similar problem involving remainders and squaring a number.

Output: invalid



Example 6:
Problem:
What does the notation $\\vec{{B}}$ mean in the context of vectors?

Output: invalid


Example 7:
Problem:
Is there a quick way to multiply 59 and 61? If so, explain the method

Output: invalid



Example 8:
Problem:
None\n\n  (Note: There is only one problem in the given forum post.)

Output: invalid



Example 9:
Problem:
If $a+b=31$ and $ab=240$, find the sum of the reciprocals of $a$ and $b$.

Output: not invalid


Example 10:
Problem:
What is the value of $35461^54593428$ \\mod 11$?

Output: not invalid


Now here is the problem you need to extract the answer from.


Problem:
{problem}

Output:
\end{lstlisting}
\end{tcolorbox}

\subsection{Multiple Choice Problem Classification}
\label{sec:multiple_choice_problem_classification}

\begin{tcolorbox}[breakable,width=\textwidth,colback=white,colframe=NvidiaGreen,title={\centering \large  \textbf{Prompt: Multiple Choice Problem Classification}}]
\footnotesize                  
\begin{lstlisting}[
breaklines=true, postbreak={},breakindent=0pt, 
label={lst:math-prompt-mcq}]
I will provide a math problem, and you need to determine whether it is a multiple-choice problem.
Respond only with 'mcq' if the problem meets the criteria, and 'not mcq' otherwise.


A multiple-choice problem must satisfy all of the following conditions:

1. The problem explicitly presents a set of answer choices to select from.
2. The problem asks for a final answer rather than requiring a proof, justification, or explanation.
3. The problem has at least one correct answer among the given choices.
If the problem does not include answer choices, even if it has a numerical answer, it should be classified as 'not mcq'.


Here are a few examples.

Example 1

Problem:
Simplify the expression \\(\\frac{{{{2}} \\sqrt{{6}}}}{{\\sqrt{{2}}}} + \\sqrt{{3}} + \\sqrt{{5}}\\) and choose the correct option from the following:\n\n
A. \\(\\sqrt{{2}} + \\sqrt{{3}} - \\sqrt{{5}}\\)\n\n
B. \\(4 - \\sqrt{{2}} - \\sqrt{{3}}\\)\n\n
C. \\(\\sqrt{{2}} + \\sqrt{{3}} + \\sqrt{{6}} - 5\\)\n\n
D. \\(\\frac{{1}}{{2}} (\\sqrt{{2}} + \\sqrt{{5}} - \\sqrt{{3}})\\)\n\n
E. \\(\\frac{{1}}{{3}} (\\sqrt{{3}} + \\sqrt{{5}} - \\sqrt{{2}})\\)

Output: mcq



Example 2

Problem:
Write first several terms of a geometric progression in which the difference between the third and first terms is equal to 9, and that between the fifth and third terms equal 36.

Output: not mcq



Example 3

Problem:
Solve the following equations: $\frac{{\sin(60^\circ+x)+\sin(60^\circ-x)}}{{2}} =\frac{{\tan x}}{{(1+\tan^2 x)^2}}+\frac{{\cot x}}{{(1+\cot^2 x)^2}}$

Output: not mcq




Example 4

Problem:
Simplify the expression \\(\\log{{\\frac{{a}}{{b}}}} + \\log{{\\frac{{b}}{{c}}}} + \\log{{\\frac{{c}}{{d}}}} - \\log{{\\frac{{ay}}{{dx}}}}\\).
Choose from the following options:\n\\[\n
\\textbf{{(A)}}\\ \\log{{\\frac{{y}}{{x}}}}\\qquad
\\textbf{{(B)}}\\ \\log{{\\frac{{x}}{{y}}}}\\qquad
\\textbf{{(C)}}\\ 1\\qquad
\\textbf{{(D)}}\\ 0\\qquad
\\textbf{{(E)}}\\ \\log{{\\frac{{a^2y}}{{d^2x}}}}\n\\]

Output: mcq



Example 5:
Problem:
What is the maximum possible magnitude of the difference between two vectors? Choose from the following options and provide reasoning:
A. The magnitude of one of the vectors.
B. The magnitude of both vectors.
C. The magnitude of their sum.
D. Their scalar product.

Output: mcq



Example 6:
Problem:
Compare the numbers $a$ and $b$: $a=3(\log 7-\log 5),\ b=2\left(\frac{{1}}{{2}}\log 9-\frac{{1}}{{3}}\log 8\right)$

Output: not mcq


Example 7:
Problem:
Which of the two numbers $31^{{11}}$ and $17^{{14}}$ is greater?

Output: not mcq


Example 8:
Problem:
Let $ABCD$ be a rectangle and $E$ the reflection of $A$ with respect to the diagonal $BD$. If $EB = EC$, what is the ratio $\frac{{AD}}{{AB}}

Output: not mcq



Now here is the problem you need to extract the answer from.


Problem:
{problem}

Output:
\end{lstlisting}
\end{tcolorbox}

\subsection{Proof Problem Classification}
\label{sec:proof_problem_classification}

\begin{tcolorbox}[breakable,width=\textwidth,colback=white,colframe=NvidiaGreen,title={\centering \large  \textbf{Prompt: Proof Problem Classification}}]
\footnotesize                  
\begin{lstlisting}[
breaklines=true, postbreak={},breakindent=0pt, 
label={lst:math-prompt-proof}]
I will give you a math problem and ask to identify if it's a "proof" problem.
Respond only with "proof" if it is a proof problem, and "not proof" if it is not.

Consider the following characteristics of proof problems:

1. They often use phrases like "prove that", "show that", or "demonstrate that".
2. They may ask to justify or explain why a statement is true.
3. They don't have a well-defined answer in the form of a number or expression.

Here are a few examples.

Example 1

Problem:
Prove the identity $a^\frac{{1}}{{2}}-\frac{{a-a^{{-2}}{{a^\frac{{1}}{{2}}-a^{{-\frac{{1}}{{2}} +\frac{{1-a^{{-2}}{{a^\frac{{1}}{{2}}+a^{{-\frac{{1}}{{2}}+\frac{{2}}{{a^\frac{{3}}{{2}}=0$

Output: proof



Example 2

Problem:
Write first several terms of a geometric progression in which the difference between the third and first terms is equal to 9, and that between the fifth and third terms equal 36.

Output: not proof



Example 3

Problem:
Solve the following equations: $\frac{{\sin(60^\circ+x)+\sin(60^\circ-x)}}{{2}} =\frac{{\tan x}}{{(1+\tan^2 x)^2}}+\frac{{\cot x}}{{(1+\cot^2 x)^2}}$

Output: not proof



Example 4

Problem:
Denoting the sums of the first $n_1$, first $n_2$ and first $n_3$ terms of an arithmetic progression by $S_1$, $S_2$ and $S_3$, respectively, show that $$\frac{{S_1}}{{n_1}}(n_2-n_3)+\frac{{S_2}}{{n_2}}(n_3-n_1)+\frac{{S_3}}{{n_3}}(n_1-n_2)=0.$$

Output: proof




Now here is the problem you need to extract the answer from.


Problem:
{problem}

Output:
\end{lstlisting}
\end{tcolorbox}

\subsection{Proof Problem Conversion}
\label{sec:proof_problem_conversion}

\begin{tcolorbox}[breakable,width=\textwidth,colback=white,colframe=NvidiaGreen,title={\centering \large  \textbf{Prompt: Proof Problem Conversion}}]
\footnotesize                  
\begin{lstlisting}[
breaklines=true, postbreak={},breakindent=0pt, 
label={lst:math-prompt-convert-proofs}]
I will give you a math problem that asks to prove something.
Your task is to create an equivalent problem that instead has some kind of numerical or expression answer that can be used to automatically grade the solution.
Make sure the new problem is at least as difficult as the original proof problem.

Here are a few examples.

Example 1

Problem:
Prove that the system \begin{{align*}}x^6 + x^3 + x^3y + y &= 147^{{157}} \\ x^3 + x^3y + y^2 + y + z^9 &= 157^{{147}}\end{{align*}} has no solutions in integers $x$ , $y$ , and $z$ .

Output:
Let $x$, $y$ and $z$ be a solution to the following system of equations \begin{{align*}}x^6 + x^3 + x^3y + y &= 147^{{157}} \\ x^3 + x^3y + y^2 + y + z^9 &= 157^{{147}}\end{{align*}}.
Calculate the sum of all possible values of $x$.



Example 2

Problem:
A triangle is called a parabolic triangle if its vertices lie on a
parabola $y = x^2$ . Prove that for every nonnegative integer $n$ , there
is an odd number $m$ and a parabolic triangle with vertices at three
distinct points with integer coordinates with area $(2^nm)^2$ .

Output:
Consider parabolic triangles whose vertices lie on $y = x^2$ with integer coordinates. Let $f(n)$ be the smallest possible value of $c$, where $(0,0)$, $(b,b^2)$, and $(c,c^2)$ are vertices of such a triangle with area exactly $(2^n)^2$, for some integer b where 0 < b < c.
Find $f(4)$.



Now here is the problem you need to modify. Only output the new problem **WITH NO** explanation or notes after it.
Again, start with the problem right away, **DO NOT** start with "Let's modify the given problem" or anything like that.

Problem:
{problem}

Output:
\end{lstlisting}
\end{tcolorbox}

\subsection{Forum Answer Extraction}
\label{sec:forum_answer_extraction}

\begin{tcolorbox}[breakable,width=\textwidth,colback=white,colframe=NvidiaGreen,title={\centering \large  \textbf{Prompt: Forum Answer Extraction}}]
\footnotesize                  
\begin{lstlisting}[
breaklines=true, postbreak={},breakindent=0pt, 
label={lst:math-prompt-extract-answers}]
I will give you a series of posts from a math-related forum that contain one or several math problems and discussion of their solutions.
I will also specify which problem I'm currently looking at (in case there are multiple).
You task is to find an answer to the problem I'm currently looking at inside the forum discussions.
The answer should be a numerical value or a mathematical expression.
If the answer is not available, output "Answer not found." in the last line of your response.
You can think before stating the final answer. The final line of your response should be "Answer: <final answer>".

Here is an example.

First forum post with problem(s):
This problem was extra credit for my math class and I haven't gotten it back yet but I'm assuming
a.) Everyone handed it in
and
b.) None of you here goes/takes/will go/take my math class

Anyways:
Suppose two of the zeroes of the following fourth-degree equation are the same and the other two zeroes are the recipricals of each other. Find a and b.

x:^4:+ax:^3:+bx:^2:+4x+4=0


It's not at all hard as it looks...a lot of work though, so I suggest orginizing as you go along.

Problem we are looking at (it might be rephrased):
Suppose two of the zeroes of the fourth-degree equation \( x^4 + ax^3 + bx^2 + 4x + 4 = 0 \) are the same and the other two zeroes are reciprocals of each other. Find \( a \) and \( b \).

Forum discussions:
Post 1:
Tare wrote:x:^4:+ax:^3:+bx:^2:+4x+4=0




Here's a shorter way:

[hide]Say the four roots are c, c, d, and 1/d.  Then the product of the four roots is the constant term of the polynomial, so c^2=4.  Then c= :pm: 2.  Similarly, from the linear term, c^2d+c^2/d+c+c=-4.  If we plug in c=2, we get d=-1,  so the roots are 2, 2, -1, -1.  So a = -(2+2-1-1)=-2 and b = 2*2+2(-1)+2(-1)+2(-1)+2(-1)+(-1)(-1)=-3.  If we plug in c=-2, we get 4d+4/d=0, so d+1/d=0.  Then a = -(-2-2+0)=4 and b=(-2)(-2)+(-2)(0)+(-2)0+1=5.  So either a=-2, b=-3 or a=4,b=5.

[/hide]

Thanks Tare for catching the mistakes.



--Dan

Post 2:
Well...it didn't specify that the solution is real and also you were supposed to get a and b...

Output:
Seems that there is an answer at the end of the first post. Since none of the other posts contradicts it, we can assume that the answer is correct.
Answer: a = -2, b = -3 or a = 4, b = 5



Now here are the posts from the forum that I'm currently looking at. Please find the answer to the problem.
Don't forget to say "Answer not found." if the answer is not available.

First forum post with problem(s):
{forum_post}

Problem we are looking at (it might be rephrased):
{problem}

Forum discussions:
{forum_discussions}

Output:
\end{lstlisting}
\end{tcolorbox}

\subsection{Forum Problem Extraction}
\label{sec:forum_problem_extraction}

\begin{tcolorbox}[breakable,width=\textwidth,colback=white,colframe=NvidiaGreen,title={\centering \large  \textbf{Prompt: Forum Problem Extraction}}]
\footnotesize                  
\begin{lstlisting}[
breaklines=true, postbreak={},breakindent=0pt, 
label={lst:math-prompt-extract-problems}]
I will give you a post from a math-related forum that might contain one or several math problems.
Your task is to extract all problems or state that none are available.

Here are some guidelines you should follow
- If no problems are available, output "No problems identified."
- For each problem found, use the following format:
Problem 1: <problem statement>

Problem 2: <problem statement>

...

- For each math problem you identify, make sure to rephrase it such that it's stated clearly and concisely.
Remove any redundant context, personal commentary, anecdotes, or unrelated information.
But make sure not to change the meaning of the problem and keep all necessary mathematical or technical details.
- If multiple problems that you extract are related, make sure to include all the context in each problem statement
as they will be looked at independently.

Here are a few examples.


Example 1

Forum post:
Countdown:

What is the remainder of 8^6+7^7+6^8 is divided by 5?

no calculator of course, paper isn't needed either, but sure.

Output:
Problem 1: What is the remainder of $8^6+7^7+6^8$ when divided by 5?



Example 2

Forum post:
Question 1:
A tetrahedron has four vertices. We can label each vertex by one of the four digits: $1, 2, 3, 4$. How many non-congruent ways are there to assign a different digit to each vertex of a tetrahedron? Tetrahedra are considered congruent through rotation. Reflections are considered different.

I'm wondering how I could approach a problem like this. I started off with $4!$ and then divided by $4$ to take out the rotation aspect. Now I am stuck.

Note: I'd rather not do case work because I'm sure the test writers could have easily used an icosahedron, or something equally lengthy.

Another Question along the same lines:

How many ways to color a cube using 6 colors, where each face has a unique color?

Thanks

Output:
Problem 1: How many non-congruent ways are there to assign a different digit to each vertex of a tetrahedron? Tetrahedra are considered congruent through rotation. Reflections are considered different.
Problem 2: How many ways can a cube be colored using 6 colors, where each face has a unique color?



Example 3

Forum post:
Yes! I completely agree with what you said. It's been a tough transition for me too, but we'll figure it out.

Output:
No problems identified



Example 4

Forum post:
Billy Bob has fourteen different pairs of socks in his drawer. They are just thrown around randomly in the drawer. Billy Bob once woke up in a hurry and had to get his socks quickly.
Without switching the light on, he pulled out enough socks to know that he had at least one pair, and then he ran out of the room. How many socks did Billy Bob pull out


Output:
Problem 1: From a drawer containing 14 different pairs of socks, how many socks must be pulled out randomly to ensure at least one matching pair?



Please analyze the following forum post and extract all math problems. Here are the guidelines one more time for your reference
- If no problems are available, output "No problems identified."
- For each problem found, use the following format:
Problem 1: <problem statement>

Problem 2: <problem statement>

...

- For each math problem you identify, make sure to rephrase it such that it's stated clearly and concisely.
Remove any redundant context, personal commentary, anecdotes, or unrelated information.
But make sure not to change the meaning of the problem and keep all necessary mathematical or technical details.
- If multiple problems that you extract are related, make sure to include all the context in each problem statement
as they will be looked at independently.


Forum post:
{forum_post}

Output:
\end{lstlisting}
\end{tcolorbox}

\section{TIR Data Generation Prompts}

\subsection{Stage-0 TIR Data Generation Prompt}
\label{sec:tir_instruction}

\begin{tcolorbox}[breakable,width=\textwidth,colback=white,colframe=NvidiaGreen,title={\centering \large  \textbf{TIR Inference Prompt for Stage-0 Data Generation}}]
\footnotesize                  
\begin{lstlisting}[
breaklines=true, postbreak={},breakindent=0pt, 
label={lst:math-prompt-tir-detailed}]
You are a math problem solver that uses Python code as an integral part of your reasoning.
In your solution you MUST strictly follow these instructions:
1. For each step requiring complex calculation write Python code.
2. For Python code use the following template:
```python
# Your Python code
```
3. Put the final answer within \boxed{{}}.

Please reason step by step, and put your final answer within \\boxed{{}}.

user: |-
Solve the following math problem using Python code for the calculations.

{problem}
\end{lstlisting}
\end{tcolorbox}

\subsection{TIR Novelty Evaluation}
\label{sec:TIR_usage_classification}

\begin{tcolorbox}[breakable,width=\textwidth,colback=white,colframe=NvidiaGreen,title={\centering \large  \textbf{Prompt to evaluate TIR novelty}}]
\footnotesize                  
\begin{lstlisting}[
breaklines=true, postbreak={},breakindent=0pt, 
label={lst:math-prompt-tir-novelty}]
You will be given a fragment of a solution to a math problem that includes a Python code block.
Your task is to determine the purpose of this Python code block in the solution fragment.
In your assessment, you MUST follow these guidelines:

1. Classification:
- Verification: Python code is used to verify the correctness of the previous manual calculations or to confirm some results. E.g. if the result of the code execution exists in the solution above, it is definitely a verification.
- Novel Calculation: Otherwise, if the result of code execution is not present in ANY FORM in the solution above, it is a novel calculation.

If you are unsure about the classification of specific code block, you MUST label it as Verification!

2. Output Format:
- Your response MUST follow this exact format (without extra commentary or text):
    ```
    Reasoning: <a couple of sentences explaining your rationale>
    Judgement: <Verification or Novel Calculation>
    ```

---

**EXAMPLES**

1.
"""
Solution:
<Some text reasoning without code>

Wait, so the answer is 143? Let me verify this with the pow function.

```python
# Compute 7^999 mod 1000 using pow function
print(pow(7, 999, 1000))  # Should print 143
```
```output
143
```
So the answer is \\boxed{{143}}.
"""
```
Reasoning: This is for sure a verification, because the result of the code execution is present in the solution above. Moreover, comment in the code block explicitly states that it should print 143 which means that the result is known in advance.
Judgement: Verification
```
---

2.
"""
Solution:
<Some text reasoning without code>

Therefore, let's proceed to compute P^5. I can use Python for this calculation to ensure accuracy.

First, let's define the transition matrix P as a numpy array, then compute P^5, then extract the relevant entry, and divide by 3.

Let me import numpy and do the calculations.
```python
import numpy as np

# Define the transition matrix P
P = np.array([
    [0, 1, 0, 0],
    [1/3, 0, 2/3, 0],
    [0, 2/3, 0, 1/3],
    [0, 0, 1, 0]
])

# Compute P^5
P5 = np.linalg.matrix_power(P, 5)

# The initial state is state 0, so the distribution after 5 steps is [1, 0, 0, 0] @ P5
# But since P5 is the transition matrix after 5 steps, the entry (0,1) is the probability of going from 0 to 1 in 5 steps.
# However, to get the distribution, we need to multiply the initial distribution by P5.
initial_distribution = np.array([1, 0, 0, 0])
distribution_after_5_steps = initial_distribution @ P5

# The probability mass at state 1 (distance 1) after 5 steps
prob_mass_at_1 = distribution_after_5_steps[1]

# Since state 1 corresponds to 3 vertices (B, D, E), the probability of being at B is this mass divided by 3
prob_at_B = prob_mass_at_1 / 3

prob_at_B
```
```output
np.float64(0.25102880658436205)
```
...
"""
```
Reasoning: The solution fragment describes algorithmic steps to calculate the probability and the code block executes these steps. The result of the code execution is not present in the solution above in any form. Therefore, this is a novel calculation.
Judgement: Novel Calculation
```
---

3.
"""
Solution:
<Some text reasoning without code>

Compute C(51, 5):

51! / (5! * 46!) = ?

But maybe I should calculate it using Python to be accurate.
```python
import math
math.comb(51, 5)
```
```output
2349060
```
...
"""
```
Reasoning: The solution fragment describes the calculation of a combinatorial expression and the code block executes this calculation. The result of the code execution is not present in the solution above in any form. Therefore, this is a novel calculation.
Judgement: Novel Calculation
```
---

4.
"""
Solution:
<Some text reasoning without code>

But let's compute these values in Python.
```python
import math

# Given dimensions
R = 4  # feet
H = 12  # feet
h = 9  # feet from the tip, so remaining height
r = (h / H) * R  # since r/R = h/H

# Original volume
V_original = (1/3) * math.pi * R**2 * H

# Remaining volume
V_remaining = (1/3) * math.pi * r**2 * h

# Volume poured out
V_poured = V_original - V_remaining

V_poured
```
```output
116.23892818282235
```
When I computed the volume manually, I obtained \( 37\pi \) cubic feet. Approximating this as $$
37 * 3.14159 \approx 116.23
$$, it closely matches the Python result of approximately 116.2389. Therefore, the result appears to be correct.

...
"""
```
Reasoning: The rationale right after the code block states that the manual calculation (that happened before the code block) matches the Python result. Therefore, code block verifies the previous manual calculations. So, this is a verification.
Judgement: Verification
```
---

**REMINDER**
Focus only on the Python code block in the provided fragment and classify it as either Verification or Novel Calculation based on whether its output appears in the solution text before the code.
---

**YOUR TASK**

Solution fragment: {fragment}
\end{lstlisting}
\end{tcolorbox}

\subsection{TIR Significance Evaluation}
\label{sec:TIR_significance_evaluation}

\begin{tcolorbox}[breakable,width=\textwidth,colback=white,colframe=NvidiaGreen,title={\centering \large  \textbf{Prompt to evaluate TIR significance}}]
\footnotesize                  
\begin{lstlisting}[
breaklines=true, postbreak={},breakindent=0pt, 
label={lst:math-prompt-tir-significance}]
You will be given a fragment of a solution to a math problem that includes a Python code block.
Your task is to evaluate the significance of this Python code in solving the math problem.
In your assessment, you MUST follow these guidelines:

1. Classification:
Evaluate the significance of the code's contribution by categorizing it into one of three levels:

- Trivial: The code performs calculations that could easily be done manually without significant effort (e.g., solving simple equations, doing arithmetic, applying formulas to known variables). The code usage provides no meaningful or minor advantage over manual calculation.

- Moderate: The code performs calculations that would be tedious, error-prone, or time-consuming to do manually, but still technically possible (e.g., matrix operations, numerical integration of standard functions, solving systems of equations). The code usage provides efficiency but isn't essential.

- Significant: The code performs calculations that would be practically impossible or extremely difficult to do manually (e.g., brute-forcing combinatorial problems, complex simulations, solving complex differential equations, high-dimensional optimization). The code usage creates a crucial shortcut that fundamentally enables the solution.

2. Output Format:
- Your response MUST follow this exact format (without extra commentary or text):
    ```
    Reasoning: <a couple of sentences explaining your rationale>
    Significance: <Trivial, Moderate, or Significant>
    ```

---

**EXAMPLES**

1.
"""
Let's find the roots of the quadratic equation: 3x^2 - 5x + 2 = 0

```python
import numpy as np
from sympy import symbols, solve, Eq

x = symbols('x')
equation = 3*x**2 - 5*x + 2
solutions = solve(equation, x)
print(solutions)
```
```output
[2/3, 1]
```

So the solutions are x = 2/3 and x = 1.
"""
```
Reasoning: This code simply solves a basic quadratic equation that could easily be solved manually using the quadratic formula or factoring. Finding roots of a quadratic equation with small integer coefficients is a standard calculation that requires minimal effort by hand.
Significance: Trivial
```
---

2.
"""
To solve this system of 4 linear equations with 4 unknowns:
3x + 2y - z + 2w = 10
x - y + 2z - w = -1
2x + y + z + 3w = 12
x + 3y - z - w = 5

I'll use Python to solve this system using matrices.

```python
import numpy as np
from scipy import linalg

# Define coefficient matrix
A = np.array([
    [3, 2, -1, 2],
    [1, -1, 2, -1],
    [2, 1, 1, 3],
    [1, 3, -1, -1]
])

# Define constants vector
b = np.array([10, -1, 12, 5])

# Solve the system
solution = linalg.solve(A, b)
print("x =", solution[0])
print("y =", solution[1])
print("z =", solution[2])
print("w =", solution[3])
```
```output
x = 0.64
y = 2.7
z = 1.6
w = 2.14
```

Therefore, the solution is x = 0.64, y = 2.7, z = 1.6, and w = 2.14.
"""
```
Reasoning: This code solves a system of 4 linear equations with 4 unknowns. While this could be solved manually using Gaussian elimination or Cramer's rule, it would be tedious and error-prone. The system is complex enough that computational assistance provides significant efficiency but doesn't enable something impossible.
Significance: Moderate
```
---

3.
"""
For this traveling salesman problem with 11 cities, where the distances between cities are given in the distance matrix below, I need to find the shortest possible route that visits each city exactly once and returns to the starting city.

```python
import numpy as np
from itertools import permutations
import time

# Distance matrix (11x11) between cities
distances = np.array([
    [0, 29, 82, 46, 68, 52, 72, 42, 51, 55, 29],
    [29, 0, 55, 46, 42, 43, 43, 23, 23, 31, 41],
    [82, 55, 0, 68, 46, 55, 23, 43, 41, 29, 79],
    [46, 46, 68, 0, 82, 15, 72, 31, 62, 42, 21],
    [68, 42, 46, 82, 0, 74, 23, 52, 21, 46, 82],
    [52, 43, 55, 15, 74, 0, 61, 23, 55, 31, 33],
    [72, 43, 23, 72, 23, 61, 0, 42, 23, 31, 77],
    [42, 23, 43, 31, 52, 23, 42, 0, 33, 15, 37],
    [51, 23, 41, 62, 21, 55, 23, 33, 0, 29, 62],
    [55, 31, 29, 42, 46, 31, 31, 15, 29, 0, 51],
    [29, 41, 79, 21, 82, 33, 77, 37, 62, 51, 0],
])

# Brute force approach to solve TSP
def tsp_exact(distances):
    n = len(distances)
    cities = list(range(1, n))  # Start from city 0
    min_length = float('inf')
    best_route = None
    
    start_time = time.time()
    count = 0
    
    # Try all possible permutations of cities (excluding starting city)
    for perm in permutations(cities):
        route = (0,) + perm + (0,)  # Complete route starting and ending at city 0
        length = sum(distances[route[i]][route[i+1]] for i in range(len(route)-1))
        
        count += 1
        if length < min_length:
            min_length = length
            best_route = route
    
    end_time = time.time()
    return best_route, min_length, count, end_time - start_time

# Solve the TSP problem
best_route, min_length, permutations_tried, time_taken = tsp_exact(distances)

print(f"Best route: {{best_route}}")
print(f"Minimum distance: {{min_length}}")
print(f"Permutations evaluated: {{permutations_tried:,}}")
print(f"Time taken: {{time_taken:.2f}} seconds")
```
```output
Best route: (0, 1, 8, 4, 6, 2, 9, 7, 5, 3, 10, 0)
Minimum distance: 251
Permutations evaluated: 3,628,800
Time taken: 5.77 seconds
```

Therefore, the optimal route has a total distance of 291 units.
"""
```
Reasoning: This code solves a Traveling Salesman Problem with 11 cities by evaluating over 3.6M permutations - a computation that would be absolutely impossible to do manually. The brute-force approach here creates a crucial shortcut to the solution that would be practically unattainable through manual calculation, even with significant time investment.
Significance: Significant
```
---

4.
"""
To find all integer solutions to the Diophantine equation 17x + 23y = 3284 where both x and y are non-negative, I'll implement search in Python.

```python
def find_solutions(a, b, c):
    solutions = []
    
    # Find the maximum possible value of x
    max_x = c // a
    
    # Check all possible values of x from 0 to max_x
    for x in range(max_x + 1):
        # Calculate the corresponding y value
        remaining = c - a * x
        
        # If remaining is divisible by b and the result is non-negative,
        # we have a valid solution
        if remaining >= 0 and remaining % b == 0:
            y = remaining // b
            solutions.append((x, y))
    
    return solutions

# Given equation: 17x + 23y = 3284
a, b, c = 17, 23, 3284
solutions = find_solutions(a, b, c)

print(f"Solutions to {{a}}x + {{b}}y = {{c}}:")
for x, y in solutions:
    print(f"x = {{x}}, y = {{y}}")
    # Verify the solution
    print(f"Verification: {{a}}*{{x}} + {{b}}*{{y}} = {{a*x + b*y}}")
    print()
```
```output
Solutions to 17x + 23y = 3284:
x = 20, y = 128
Verification: 17*20 + 23*128 = 3284

x = 43, y = 111
Verification: 17*43 + 23*111 = 3284

x = 66, y = 94
Verification: 17*66 + 23*94 = 3284

x = 89, y = 77
Verification: 17*89 + 23*77 = 3284

x = 112, y = 60
Verification: 17*112 + 23*60 = 3284

x = 135, y = 43
Verification: 17*135 + 23*43 = 3284

x = 158, y = 26
Verification: 17*158 + 23*26 = 3284

x = 181, y = 9
Verification: 17*181 + 23*9 = 3284

```
So the integer solutions to the Diophantine equation are x = 11, y = 1.
"""
```
Reasoning: This code finds all integer solutions to a Diophantine equation by iterating through possible values of x and calculating the corresponding y. While this could be done manually, the exhaustive search for non-negative integer solutions is tedious and error-prone. The computational approach reduces the effort and simplifies the solution process, making it more efficient. Thus it provides a moderate level of significance.
Significance: Moderate
```
---

5.
"""
To verify my hypothesis, I need to find the probability of getting at least 3 heads in 10 coin flips. I'll calculate this using the binomial distribution.

```python
import math

def binomial_probability(n, k, p):
    # Calculate the probability of k successes in n trials
    # with probability p of success on a single trial
    combinations = math.comb(n, k)
    return combinations * (p ** k) * ((1-p) ** (n-k))

# Calculate P(X \geq 3) when flipping a fair coin 10 times
p_at_least_3 = sum(binomial_probability(10, k, 0.5) for k in range(3, 11))

print(f"P(X \geq 3) = {{p_at_least_3:.6f}}")
print(f"Percentage: {{p_at_least_3 * 100:.2f}}%")
```
```output
P(X \geq 3) = 0.945312
Percentage: 94.53%
```

So the probability of getting at least 3 heads in 10 coin flips is approximately 94.53%.
"""
```
Reasoning: This code calculates a probability using the binomial distribution formula. While the calculation involves combinations and powers, the mathematical concept is straightforward and could be calculated manually by explicitly writing and reducing the terms. The code provides a minor computational convenience but doesn't fundamentally change the nature of the solution process, making it a trivial use of Python code.
Significance: Trivial
```
---

**REMINDER**
When evaluating significance, consider:
1. Could this calculation reasonably be done by hand? If yes, how difficult would it be?
2. Does the code enable a solution approach that would otherwise be impractical?
3. Is the computational advantage merely convenience, or is it essential to the solution?

Remember to classify as Trivial, Moderate, or Significant based on these considerations.
---

**YOUR TASK**

Solution fragment: {fragment}
\end{lstlisting}
\end{tcolorbox}

\section{Prompts for Different Inference Modes}
\label{sec:inference-prompts}

\subsection{CoT Inference}
\label{sec:generic_math_solve}

\begin{tcolorbox}[breakable,width=\textwidth,colback=white,colframe=NvidiaGreen,title={\centering \large  \textbf{CoT Inference Prompt}}]
\footnotesize                  
\begin{lstlisting}[
breaklines=true, postbreak={},breakindent=0pt, 
label={lst:math-prompt-cot}]
Solve the following math problem. Make sure to put the answer (and only answer) inside \boxed{{}}.

{problem}
\end{lstlisting}
\end{tcolorbox}

\subsection{TIR Inference}
\label{sec:tir_math-solve}

\begin{tcolorbox}[breakable,width=\textwidth,colback=white,colframe=NvidiaGreen,title={\centering \large  \textbf{TIR Inference Prompt}}]
\footnotesize                  
\begin{lstlisting}[
breaklines=true, postbreak={},breakindent=0pt, 
label={lst:math-prompt-tir}]
Solve the following math problem, integrating natural language reasoning with Python code executions.
You may perform up to {total_code_executions} Python code calls to assist your reasoning.
Make sure to put the answer (and only answer) inside \boxed{{}}.

{problem}
\end{lstlisting}
\end{tcolorbox}

\subsection{GenSelect Inference}
\label{sec:math_genrm_selection}

\begin{tcolorbox}[breakable,width=\textwidth,colback=white,colframe=NvidiaGreen,title={\centering \large  \textbf{GenSelect Inference Prompt}}]
\footnotesize                  
\begin{lstlisting}[
breaklines=true, postbreak={},breakindent=0pt, 
label={lst:math-prompt-genrm}]
You will be given a challenging math problem followed by {num_solutions} solutions. Your task is to systematically analyze these solutions to identify the most mathematically sound approach.

Input Format:
Problem: A complex mathematical word problem at advanced high school or college level
Solutions: Detailed solutions indexed 0-{max_idx}, each concluding with an answer in \boxed{{}} notation

YOUR TASK

Problem: {problem}

Solutions:
{solutions}

Evaluation Process:

1. Initial Screening
- Group solutions by their final answers
- Identify and explain mathematical contradictions between different answers
- Eliminate solutions with clear mathematical errors

2. Detailed Analysis
For remaining solutions, evaluate:
- Mathematical precision and accuracy
- Logical progression of steps
- Completeness of mathematical reasoning
- Proper use of mathematical notation, including \boxed{{}}
- Handling of edge cases or special conditions
- For solutions containing and addressing errors, evaluate the error identification and correction methodology.

3. Solution Comparison
Compare viable solutions based on:
- Efficiency of approach
- Clarity of mathematical reasoning
- Sophistication of method
- Robustness of solution (works for all cases)

Your response should include:
1. Brief analysis of conflicting answers
2. Detailed evaluation of mathematically sound solutions
3. Justification for eliminating incorrect solutions
4. Clear explanation for selecting the best approach

End your evaluation with exactly:
Judgment: [IDX]
where IDX is the index 0-{max_idx} of the best solution.
\end{lstlisting}
\end{tcolorbox}

\section{Prompts for GenSelect Data Preparation}
\subsection{Re-generating Comparison Summary}
\label{sec:gen_summarization}

\begin{tcolorbox}[breakable,width=\textwidth,colback=white,colframe=NvidiaGreen,title={\centering \large  \textbf{Prompt to re-generate comparison summary}}]
\footnotesize                  
\begin{lstlisting}[
breaklines=true, postbreak={},breakindent=0pt, 
label={lst:math-prompt-genrm-summary}]
I will give you a critique's response which compares multiple solutions of a math problem. This response will explore different approaches, making mistakes along the way, correcting them, switching around and so on. But eventually the critique converges on its ranking of the different solutions and ends with a string "Judgment: " followed by a number corresponding to the index of the best solution. Your task is to write a clean version of the final critique without all the exploration and make sure to put the final judgment at the end. Please make sure to include the details related to arriving at the final judgment but not the tangential details.

Problem:
{problem}


Solutions:
{solutions}


Critique:
{generation}


Now write a clean version of the final critique without all the exploration. Make sure to put in:
- The details related to arriving at the final judgment.
- Clearly writing the final judgment at the end via "Judgment: " followed by the index of the best solution. 
\end{lstlisting}
\end{tcolorbox}

\subsection{Re-generating Solution Summary}
\label{sec:cleaned_solution_summarization}

\begin{tcolorbox}[breakable,width=\textwidth,colback=white,colframe=NvidiaGreen,title={\centering \large  \textbf{Prompt to re-generate solution summary}}]
\footnotesize                  
\begin{lstlisting}[
breaklines=true, postbreak={},breakindent=0pt, 
label={lst:math-prompt-solution-summary}]
I will give you a math problem and a long solution to that problem exploring different approaches, making mistakes along the way, correcting them, switching around and so on. But eventually that solution gets to the right approach and solves the problem. Your task is to write a clean version of the final correct solution without all the exploration. Cover all the details of the final solution.

Problem:
{problem}

Solution:
{generation}

Now write a clean version of the final correct solution without all the exploration but cover all the details of the final solution.
\end{lstlisting}
\end{tcolorbox}

\section{Example of Re-generated Summary}
\label{sec:summary_comparison}

\begin{tcolorbox}[colback=white, colframe=NvidiaGreen, title=\textbf{Problem}]
Call a 9-digit number a cassowary if it uses each of the digits 1 through 9 exactly once. Compute the number of cassowaries that are prime.
\end{tcolorbox}

\begin{tcolorbox}[colback=white, colframe=NvidiaGreen, title=\textbf{Deepseek R1 Response}]
<think> \\ Okay, so I need to figure out how many 9-digit numbers, which use each digit from 1 to 9 exactly once, are prime. These numbers are called cassowaries. Hmm, let's start by understanding the problem step by step.\\First, a cassowary is a 9-digit pandigital number, meaning it uses each digit 1 through 9 once. So, numbers like 123456789, but any permutation of those digits. The question is asking how many of these are prime numbers.\\Alright, so the main task is to compute the count of prime numbers among all 9-digit pandigitals. Let me think about the properties of prime numbers and pandigital numbers.\\First, let's recall that a prime number is a number greater than 1 that has no positive divisors other than 1 and itself. For a number to be prime, it can't be even (except for 2), and it can't end with 5. Also, there are some divisibility rules that might help here.\\Since we're dealing with 9-digit numbers using all digits 1-9, let's check the sum of the digits. The digits are 1+2+3+4+5+6+7+8+9 = 45. Wait, 45 is divisible by 9, right? So any number made with all these digits will have a digit sum of 45, which is divisible by 9. Therefore, any such number is divisible by 9. But if a number is divisible by 9, it can't be prime unless the number itself is 9, which it's not here because we're dealing with 9-digit numbers. So, does this mean that all cassowaries are divisible by 9 and hence composite?\\If that's the case, then there are zero prime cassowaries. But let me verify this again. The divisibility rule for 9 is that if the sum of the digits is divisible by 9, then the number itself is divisible by 9. Since 45 is divisible by 9, every permutation of the digits 1-9 will result in a number divisible by 9. Therefore, all cassowaries are divisible by 9, which means they can't be prime. So the answer should be zero.\\Wait, but let me make sure I'm not missing something. For example, could the number 9 itself be considered? But the problem specifies a 9-digit number, so 9 is a single-digit number. So all cassowaries are 9-digit numbers, which are divisible by 9, hence composite. Therefore, there are no primes in the set of cassowaries. So the answer is 0.\\But just to be thorough, let me check if there's any exception. For example, is 9 a prime? No, 9 is 3 squared. So even if there was a 1-digit cassowary (which there isn't, since cassowary is defined as 9-digit), it's not prime. So, yeah, all 9-digit pandigitals using 1-9 are divisible by 9, hence composite. Therefore, the number of prime cassowaries is zero. \\
\textbf{\textcolor{red}{</think> \boxed{0}}}
\end{tcolorbox}

\begin{tcolorbox}[colback=white, colframe=NvidiaGreen, title=\textbf{Qwen2.5-32b-instruct Solution Summary}]
To determine the number of 9-digit prime numbers (cassowaries) that use each digit from 1 to 9 exactly once, we need to analyze the properties of these numbers.

1. **Sum of the Digits**: The sum of the digits from 1 to 9 is:   1 + 2 + 3 + 4 + 5 + 6 + 7 + 8 + 9 = 45.  

2. **Divisibility by 9**: A number is divisible by 9 if the sum of its digits is divisible by 9. Since the sum of the digits is 45, which is divisible by 9, every 9-digit number formed using the digits 1 through 9 will be divisible by 9. 

3. **Prime Numbers and Divisibility**: A prime number has no positive divisors other than 1 and itself. Since any 9-digit number formed using the digits 1 through 9 is divisible by 9, it cannot be a prime number (except for the number 9 itself, which is not a 9-digit number). 

4. **Conclusion**: Therefore, no 9-digit number that uses each digit from 1 to 9 exactly once can be a prime number. Thus, the number of prime cassowaries is:\boxed{0}

\end{tcolorbox}
\section{TIR Output Examples}






\subsection{Example of a warning of the remaining code executions}
\label{sec:tir_solution_code_execution_limit}
\begin{tcolorbox}[breakable,width=\textwidth,colback=white,colframe=NvidiaGreen,title={\centering \large  \textbf{TIR solution with code execution limit warning}}]
\footnotesize                  
\begin{lstlisting}[
breaklines=true, postbreak={},breakindent=0pt, 
label={lst:non-novel-calculation}]
<tool_call>
# Initialize sum
total_sum = 0

# Iterate over possible bases b from 10 to 49 inclusive
for b in range(10, 50):
    divisor = b + 7
    dividend = 9 * b + 7
    if dividend % divisor == 0:
        total_sum += b

total_sum
</tool_call>
```output
70
```
```system
Remaining code executions: 3. You will not be able to call code when you run out of executions, so use it wisely. Note that you can still continue solving the problem without code after that.
```
\end{lstlisting}
\end{tcolorbox}

\subsection{Example of a redundant code execution}
\begin{tcolorbox}[breakable,width=\textwidth,colback=white,colframe=NvidiaGreen,title={\centering \large  \textbf{Redundant code execution that does not introduce new information to the solution}}]
\footnotesize                  
\begin{lstlisting}[
breaklines=true, postbreak={},breakindent=0pt, 
label={lst:tir-non-novel-calculation}]
Since 5 is a prime number and 13 is also a prime number, their gcd should indeed be 1. Just to be thorough, let me compute it using Python.

<tool_call>

import math

# Compute gcd of 5 and 13

gcd = math.gcd(5, 13)

print(gcd)

</tool_call>

```output

1

```

The gcd is 1, so an inverse does exist. Good.
\end{lstlisting}
\end{tcolorbox}

\subsection{Example of model encountering code execution limit}
\begin{tcolorbox}[breakable,width=\textwidth,colback=white,colframe=NvidiaGreen,title={\centering \large  \textbf{Example of models reaction to reaching code execution limit}}]
\footnotesize                  
\begin{lstlisting}[
breaklines=true, postbreak={},breakindent=0pt, 
label={lst:tir-code-limit-reached}]
Let me import the necessary libraries and define the equations.

<tool_call>
import numpy as np
from scipy.optimize import fsolve

# Define the system of equations
def equations(vars):
    x, y = vars
    eq1 = 24 * np.cos(x) + 26 * np.cos(y) - 13 * np.sqrt(3) * np.sin(y) + 17.5
    eq2 = 24 * np.sin(x) + 13 * np.sqrt(3) * np.cos(y) + 26 * np.sin(y) + 7 * np.sqrt(3)/2
    return [eq1, eq2]

# Initial guess
initial_guess = [1, 1]

# Solve the system
solution = fsolve(equations, initial_guess)
solution
</tool_call>
```output
array([-0.93580304,  2.03210427])
```
```system
You have run out of code executions! You can no longer write or execute code. Now you should continue solving the problem by relying on your mathematical reasoning and analytical skills.
```
Hmm, it seems I've reached the limit of code executions. Let me check the solution provided by the code.
\end{lstlisting}
\end{tcolorbox}

\end{document}